\documentclass[sigconf]{acmart}
\AtBeginDocument{%
  }

\usepackage[utf8]{inputenc} 
\usepackage[T1]{fontenc}    
\usepackage{hyperref}       
\usepackage{url}            
\usepackage{booktabs}       
\usepackage{amsfonts}       
\usepackage{nicefrac}       
\usepackage{microtype}      
\usepackage{xcolor}         
\usepackage{enumitem}  
\usepackage{graphicx}
\usepackage{subfig}
\usepackage{wrapfig} 
\usepackage{multirow}
\usepackage{xspace}
\usepackage{diagbox}
\usepackage{svg}
\usepackage{pifont}  
\usepackage{amsmath}
\usepackage{adjustbox}
\usepackage{colortbl}
\newcommand{\name}{MAGB\xspace}

\newcommand{\first}{\cellcolor[HTML]{52B788}}
\newcommand{\second}{\cellcolor[HTML]{95D5B2}}
\newcommand{\third}{\cellcolor[HTML]{D8F3DC}}
\setcopyright{acmlicensed}
\copyrightyear{2025}
\acmYear{2025}
\acmDOI{XXXXXXX.XXXXXXX}
\acmConference[Conference acronym 'XX]{Make sure to enter the correct
  conference title from your rights confirmation email}{June 03--05,
  2025}{Woodstock, NY}
\acmISBN{978-1-4503-XXXX-X/2018/06}

\begin{document}

\title{When Graph meets Multimodal: Benchmarking and Meditating on Multimodal Attributed Graphs Learning}


\author{Hao Yan}
\affiliation{%
  \institution{Central South University}
    \city{Changsha}
  \country{China}}
\email{CSUyh1999@csu.edu.cn}

\author{Chaozhuo Li}
\affiliation{%
  \institution{Microsoft Research Asia}
    \city{Beijing}
  \country{China}}
\email{lichaozhuo1991@gmail.com}

\author{Jun Yin}
\affiliation{%
  \institution{Central South University}
      \city{Changsha}
  \country{China}}
\email{yinjun2000@csu.edu.cn}

\author{Zhigang Yu}
\affiliation{%
  \institution{Central South University}
      \city{Changsha}
  \country{China}}
\email{244711004@csu.edu.cn}

\author{Weihao Han}
\affiliation{%
  \institution{Microsoft AI}
  \city{Beijing}
  \country{China}}
\email{weihan@microsoft.com}

\author{Mingzheng Li}
\affiliation{%
  \institution{Microsoft AI}
  \city{Beijing}
  \country{China}}
\email{mingzhengli@microsoft.com}

\author{Zhengxin Zeng}
\affiliation{%
  \institution{Microsoft AI}
  \city{Beijing}
  \country{China}}
\email{zhze@microsoft.com}

\author{Hao Sun}
\affiliation{%
  \institution{Microsoft AI}
  \city{Beijing}
  \country{China}}
\email{hasun@microsoft.com}




\author{Senzhang Wang}
\affiliation{%
  \institution{Central South University}
      \city{Changsha}
  \country{China}}
\email{szwang@csu.edu.cn}

\renewcommand{\shortauthors}{Hao Yan et al.}

\begin{abstract}
Multimodal Attributed Graphs (MAGs) are ubiquitous in real-world applications, encompassing extensive knowledge through multimodal attributes attached to nodes (e.g., texts and images) and topological structure representing node interactions. Despite its potential to advance diverse research fields like social networks and e-commerce, MAG representation learning (MAGRL) remains underexplored due to the lack of standardized datasets and evaluation frameworks. In this paper, we first propose \name, a comprehensive MAG benchmark dataset, featuring curated graphs from various domains with both textual and visual attributes. Based on \name dataset, we further systematically evaluate two mainstream MAGRL paradigms: \textit{GNN-as-Predictor}, which integrates multimodal attributes via Graph Neural Networks (GNNs), and \textit{VLM-as-Predictor}, which harnesses Vision Language Models (VLMs) for zero-shot reasoning. Extensive experiments on \name reveal following critical insights: 
\textit{(i)} Modality significances fluctuate drastically with specific domain characteristics. 
\textit{(ii)} Multimodal embeddings can elevate the performance ceiling of GNNs. However, intrinsic biases among modalities may impede effective training, particularly in low-data scenarios.
\textit{(iii)} VLMs are highly effective at generating multimodal embeddings that alleviate the imbalance between textual and visual attributes.
These discoveries, which illuminate the synergy between multimodal attributes and graph topologies, contribute to reliable benchmarks, paving the way for future MAG research. The \name dataset and evaluation pipeline are publicly available at \textcolor{brown}{\url{https://github.com/sktsherlock/MAGB}}. 
\end{abstract}




\maketitle

\section{Introduction}
\label{sec:intro}
Graphs serve as powerful tools to capture relational and structural properties across various domains, including social networks \cite{netflix}, recommender systems \cite{Transformers4Rec}, and biological networks~\cite{zhu2022survey}. In many real-world scenarios, graph nodes are associated with multimodal attributes, such as texts and images, resulting in \textbf{Multimodal Attributed Graphs} (MAGs). 
MAG data structure, prevalent in social media posts pairing textual content with visual attachments or e-commerce platforms linking product descriptions with images, encompasses extensive and correlated information. MAG representation learning (MAGRL) promises significant advances in applications like social network analysis~\cite{MFTwitter}, e-commerce mining~\cite{tao2020mgat}, and recommendation systems. However, MAGRL research remains underexplored, hindered by not only the absence of standardized and publicly available benchmarks but also the limited understanding of how to jointly utilize multimodal attributes and graph topologies.

Recently, numerous foundational benchmark efforts (e.g., CS-TAG~\cite{yan2023comprehensive}, DTG~\cite{DTGB}, TEG~\cite{TEG}) have provided a wealth of datasets for the TAG domain, propelling the rapid development of the field. Leveraging these datasets, researchers have explored various novel graph-based learning paradigms, such as \textit{LLM-as-Enhancer}~\cite{TAPE,Simteg,gpROMPT} and \textit{LLM-as-Predictor}~\cite{Graphgpt,yined22025}, among others~\cite{GLEM,glbENCH}.
However, these advancements primarily concentrate on textual attributes and do not comprehensively tackle the challenges associated with multimodal information in MAGs.
In contrast to TAGs, MAGs necessitate the effective management of heterogeneous modalities, which adds complexity to feature fusion and structural reasoning.
Meanwhile, though Vision Language Models (VLMs) have demonstrated exceptional capability in multimodal learning~\cite{VItuning,Minigpt,Mini-gemini,yined22025}, the application to graph structure data is still in its infancy. 
These limitations raise fundamental questions:
\textit{How do textual and visual modalities collectively influence graph-based representation learning? How much does multimodal information contribute to improving downstream graph-related tasks?}
Addressing these challenges requires well-curated MAG datasets and systematic evaluations of different MAGRL paradigms, laying the groundwork for advanced investigations toward MAG.

\begin{figure}[!t]
    \centering
    \includegraphics[width=0.45\textwidth]{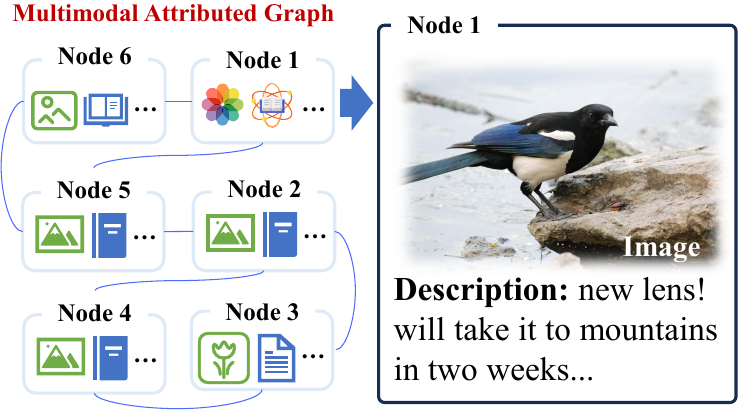}
     \caption{Illustration of a Multimodal Attributed Graph example. The left part is the MAG topological structure and the right part presents the multimodal attributes in detail.}
     \label{fig:illustrationofMAG}
     \vspace{-10pt}
\end{figure}

In this work, we introduce \textbf{MAGB}, the first benchmark dataset specifically designed for MAG representation learning (MAGRL).
MAGB dataset comprises five meticulously curated graphs from various domains, including e-commerce platforms and social networks.
As shown in Figure{\color{brown}~\ref{fig:illustrationofMAG}}, we present an example of social media MAG. 
The left part is the MAG topological structure that implies social network connections, and the right part is the multimodal attributes attached to \textit{Node 1} which consist of textual content (i.e., description) and related image. 
Specifically, the textual contents and visual images from online sources (e.g., \textit{Amazon.cn}) are preprocessed by removing missing values and anomalies, and supplemented by retrieval-based augmentation afterward.
Domain-specific concepts define entities and relationships (i.e., nodes and edges).
In e-commerce networks, nodes represent products connected by edges that indicate co-purchase relationships, whereas in social networks, nodes represent posts connected by edges that denote co-comment relationships.


Furthermore, we systematically investigate two mainstream MAGRL paradigms based on the MAGB dataset, i.e., \textit{\textbf{GNN-as-Predictor}} and \textit{\textbf{VLM-as-Predictor}}. For the former paradigm, we first adopt different kinds of feature encoder, i.e., text-only encoder, vision-only encoder, and multimodal encoder, to extract semantic information within node attributes, and then utilize the generated feature embedding as the initial node feature.
Subsequently, we aggregate MAG topology and node features through a GNN predictor and explore the influence of different modality embeddings on the GNN predictor. 
On the other hand, the \textit{VLM-as-Predictor} paradigm focuses on evaluating VLMs' zero-shot capacity.
In detail, we develop two instruction strategies: \textit{Center-only}, which merely relies on the target node attributes, and \textit{Graph Retrieval Enhancement} (GRE), which incorporates neighborhood information as an auxiliary. 
These evaluated paradigms establish a standardized framework for assessing MAGRL techniques and highlight the synergy between multimodal attributes and graph topology.

By conducting extensive experiments on the MAGB dataset, we acquire several thought-provoking insights into MAG representation learning. Notably, 
\textit{(i)} Preferences for modal attributes vary by specific domains: due to the longer text descriptions of nodes on e-commerce platforms, the textual modal contributes more to downstream tasks; whereas in social networks, posts shared by users often contain high-resolution images, making the visual modal potentially more important.
\textit{(ii)} When the training samples are abundant, multimodal embeddings can significantly enhance the performance of downstream GNNs. However, in situations with insufficient training samples, modality bias within the multimodal embeddings may hinder GNNs from fully benefiting from the diverse attributes.
\textit{(iii)} Large Vision Language Models (VLMs) are highly effective in generating multimodal embeddings, significantly improving performance and alleviating the modality imbalance caused by the substantial differences in contributions between textual and visual features in the MAG dataset.
\textit{(iv)} VLMs exhibit strong zero-shot reasoning capabilities, demonstrating versatility and generalization in node classification tasks. However, the issue of uncontrollable output persists, with a significant portion of predictions potentially falling outside the true categories in certain datasets.
\textit{(v)} \textit{Graph retrieval enhancement} can improve the upper limits of different VLMs, with increased neighbor retrieval contributing positively to performance. However, simply merging textual and visual information from neighbors may not enhance VLM reasoning and could even underperform compared to single-modal retrieval approaches.
Generally, our contributions are summarized as follows: 


\setlist{nolistsep}
\begin{enumerate}[leftmargin=*,noitemsep]
    \item We propose \name, a pioneering comprehensive benchmark for MAG representation learning research. The MAGB dataset encompasses five multimodal attributed graphs from different real-world domains, along with standardized node attributes across textual and visual modalities.

    \item We develop and implement two typical MAG representation learning paradigms: \textit{GNN-as-Predictor} and \textit{VLM-as-Predictor}, to explore the effectiveness of multimodal information and the reasoning capability of leading VLMs, respectively.

    \item We conduct extensive experiments based on the \name dataset with various experimental settings, undertake in-depth analyses, and synthesize a series of insightful conclusions, to contribute a reliable MAG benchmark and foster the MAG research community. The MAGB dataset and the MAGRL paradigms are available at \textcolor{brown}{\url{https://github.com/sktsherlock/MAGB}}.
\end{enumerate} 

\section{\name: Dataset and Benchmark for MAGs} \label{sec:bechmark}

 
In this section, we first introduce basic notations throughout the paper, followed by the construction procedure of MAGB dataset. At last, we elucidate the two typical MAGRL paradigms in detail.


\subsection{Basic Notation}
In this study, we define a \textbf{Multimodal Attributed Graph (MAG)} as $\mathcal{G} = \left(\mathcal{V}, A, \{\mathcal{M}_n\}_{n \in \mathcal{V}}\right)$, where $\mathcal{V}$ represents the node set; $A\in\{0,1\}^{N\times N}$ is the adjacency matrix, if $v_i$ and $v_j$ are connected, $A_{ij}=1$, otherwise $A_{ij}=0$. Each node \( n \in \mathcal{V} \) is associated with a set of multimodal attributes, denoted as \( \mathcal{M}_n\), which indicates the set of potential modalities (e.g., textual content $\mathcal{M}_{\mathcal{T}}$, visual image $\mathcal{M}_{\mathcal{V}}$, and etc).
For a given modality \( m \in \mathcal{M}_n \), the feature embedding provided by corresponding encoder can be denoted as \( x_n^m \in \mathbb{R}^{d_m} \), where \( d_m \) is the embedding dimension for modality \( m \).  

As previous work points out \cite{yan2023comprehensive}, text attributed graph (TAG) is a special case of MAG where $\forall n\in\mathcal{V}, \mathcal{M}_n=\{\mathcal{M}_{\mathcal{T}}\}$, indicating that each node is associated exclusively to a segment of textual content. Notably, we primarily concentrate on MAGs that comply with $\forall n\in\mathcal{V}, \mathcal{M}_n=\{\mathcal{M}_{\mathcal{T}}, \mathcal{M}_{\mathcal{V}}\}$.



\subsection{{Dataset Construction}} \label{subsec:datasets}


To thoroughly investigate the role of multimodal attributes and topological structure in MAGs and the integration strategy, we systematically study various public graph datasets which originate from real-world applications. To be more specific, text attributed graph like OGB-arXiv \cite{OGB}, whose nodes represent academic articles, contains abundant textual attributes. Visual attributed graph, such as Flickr \cite{Flickr} with user-uploaded images as nodes, focuses on exploring visual information. However, on one hand, most of existing attributed graphs narrowly focus on uni-modality related researches, although multimodal attributed graphs are common in real-world scenarios. On the other hand, previously released datasets predominantly attach processed attribute embeddings as node features, which is unchangeable during the subsequent utilization and untraceable to recover original node attributes. Therefore, the two limitations strictly restrict the comprehensiveness, flexibility, and practicality of existing attributed graph datasets.


To mitigate these issues, we take proactive steps to construct a comprehensive, flexible, and practical \textbf{\underline{MAG} \underline{B}}enchmark (MAGB) dataset, which covers e-commerce platforms and social media. On the basis of Amazon website \cite{Amazon}, we extract and establish three e-commerce networks, i.e., \textit{Movies}, \textit{Toys}, and \textit{Grocery}. Each node in the e-commerce network represent a certain product, the assigned node label is consistent with its corresponding Amazon category, and the edge indicates that the two connected products are frequently clicked or purchased together. The attached node attributes are composed of product title and description (i.e., textual content) and physical product photo (i.e., visual image). For social media based MAGs, we exploit the Reddit platform and RedCaps source \cite{redcaps} to construct \textit{Reddit-\underline{S}mall} and \textit{Reddit-\underline{M}iddle} networks. Similar to e-commerce networks, each node in Reddit-S and Reddit-M represent a post along with the user-uploaded image and related statements, the node labels are assigned according to the sub-reddit class, and the linkage means that the two posts are commented by an individual user. Basic statistics of the five multimodal attributed graphs are presented in Table {\color{brown}\ref{tab:datasets}}.

To acquire a standardized MAG benchmark dataset, we subsequently preprocess the raw attributes, including textual content clean and missing image supplement. 
It is noteworthy that the average resolution of social media images tends to be higher than that of e-commercial ones, yet the average length of social media texts is much smaller. We elaborate on the finer differences between various multimodal attributed graphs, as well as the complete data processing procedure in the Appendix \ref{DatasetConstruction}.

\begin{figure}[t]
  \centering
  \setlength{\abovecaptionskip}{0.2cm}
  \includegraphics[width=0.47\textwidth]{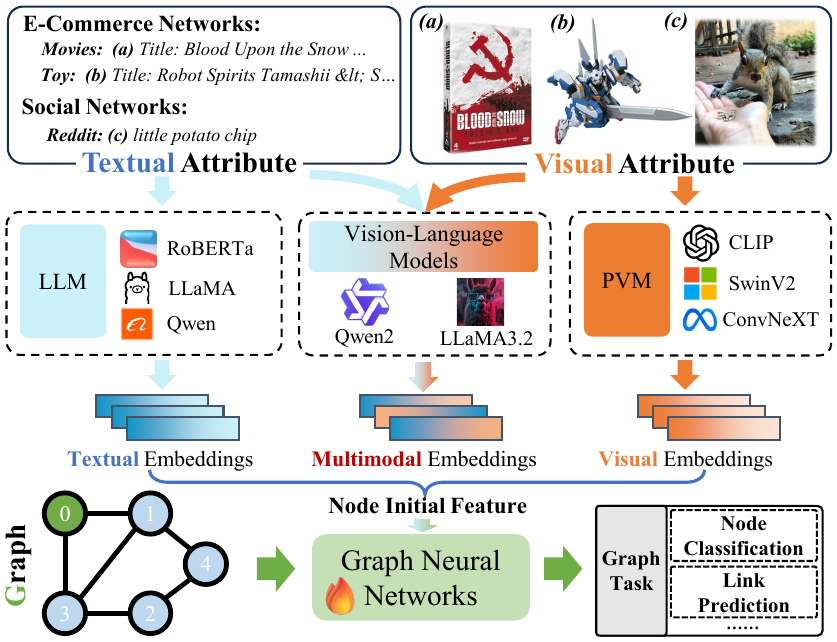}
  \caption{Overview of \textit{GNN-as-Predictor}. The attribute representations generated by modality encoders serve as the initial node features. The GNN predictor addresses various downstre am tasks based on the multimodal representations.}
  \label{fig:AEGNN}
\vspace{-0.4cm}
\end{figure}

\subsection{MAG Representation Learning} \label{subsec:baselines} 
To lay a solid foundation for MAG representation learning (MAGRL) research, we draw inspiration from the taxonomy in text attributed graph field \cite{GraphMeetLLM}, primarily exploring two typical paradigms: \textit{GNN-as-Predictor} and \textit{VLM-as-Predictor}.

\subsubsection{\textbf{GNN as Predictor}}
Based on the \textit{GNN-as-Predictor} paradigm, we target at investigating the effectiveness of different attribute modalities which embedded as initial node representations. As shown in Figure \ref{fig:AEGNN}, \textit{GNN-as-Predictor} paradigm first encodes multimodal node attributes via corresponding representation learning models, and then adopts a GNN as the task predictor to aggregate the node representations according to the MAG topological structure. In detail, \textit{GNN-as-Predictor} paradigm can be subdivide into text-only, vision-only, and multimodal versions, denoted as \textit{TE-GNN}, \textit{VE-GNN}, and \textit{ME-GNN}, respectively.
\textit{TE-GNN} refers to extracting initial node features solely based on textual attributes, where pre-trained language models (PLMs) and large language models (LLMs), such as RoBERTa \cite{Roberta} and LLaMA \cite{llamav2}, are utilized to generate textual embeddings.
\textit{VE-GNN} relies on visual attributes to generate node features for MAGs, which adopts diverse vision models, including CLIP \cite{CLIP}, ConvNext \cite{convnextv2}, and Swin Transformer \cite{swinTransformer}, to acquire visual embeddings.
\textit{ME-GNN} combines textual and visual attributes to informative multimodal knowledge, utilizing large vision language models (VLMs) \cite{llamav2} to capture cross-modality interactions and provide multimodal node representations.

\subsubsection{\textbf{VLM as Predictor}}
Considering the promising zero-shot reasoning capability of LVLMs, we aim to evaluate MAG data-mining performance by using an LVLM as the predictor. Specifically, we customize two instruction strategies for the LVLM.

\begin{figure}[t]
  \centering
  \setlength{\abovecaptionskip}{0.2cm}
  \includegraphics[width=0.475\textwidth]{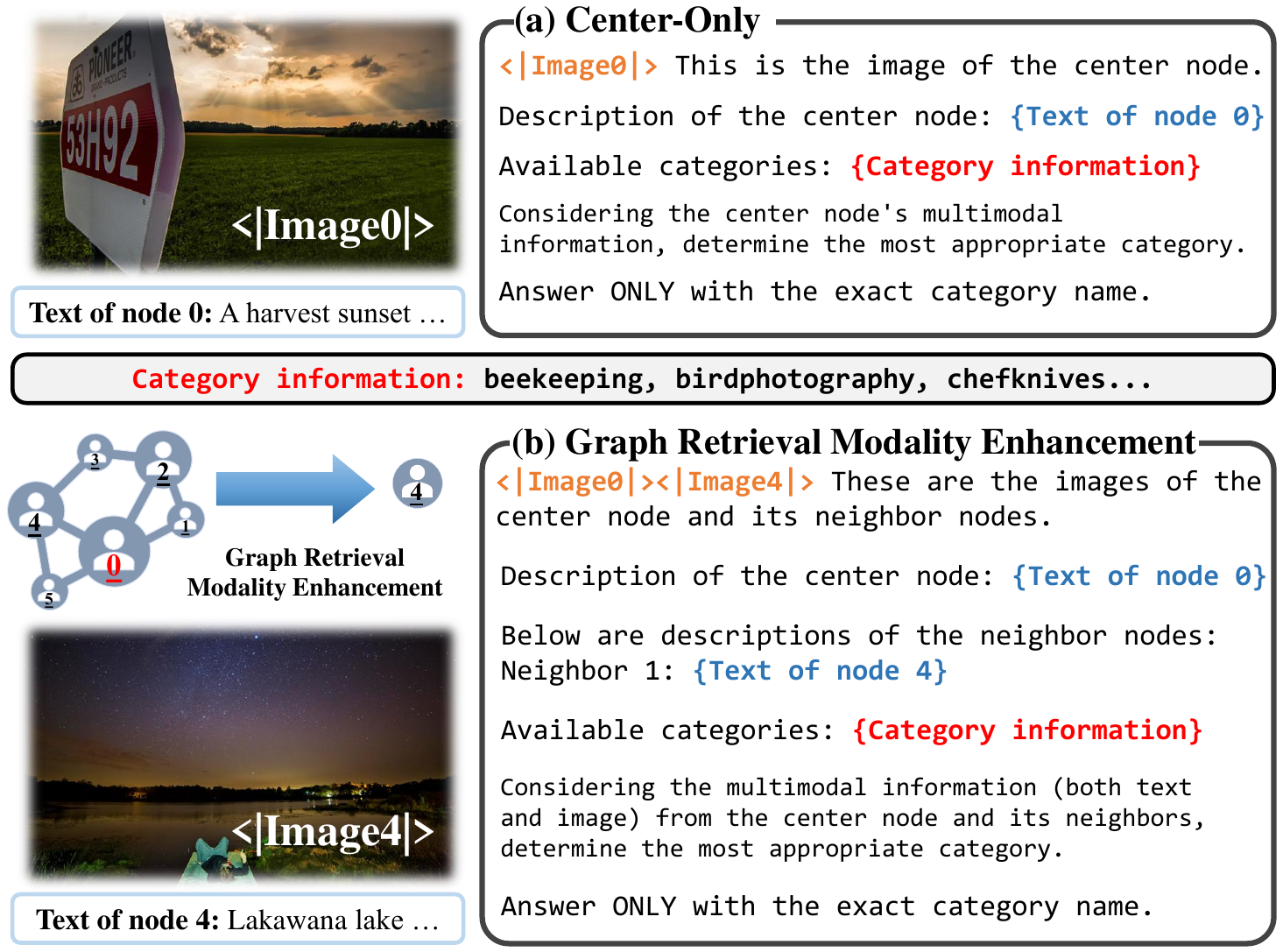}
  \caption{Overview of \textit{VLM-as-Predictor}. GRE$^k_m$ strategy first samples $k$ neighbors, and then prompts VLM with associated multimodal attributes via natural language instructions.}
  \label{fig:VLM}
\vspace{-0.4cm}
\end{figure}

\textit{Center-only} strategy directly prompts the LVLM with the multimodal attribute of the target node and requests the prediction directly. 
This prompt, as shown in Figure {\color{brown} \ref{fig:VLM}}, consists of four key components: 
the image of the center node, the textual description of the center node, the candidate set of the classification categories predefined by the data sources (e.g., \textit{Amazon.cn} or \textit{Reddit} platform), and the task-specific instruction which guides the VLM to identify the most appropriate class.



\textit{Graph Retrieval Modality Enhancement} (GRE$^k_m$) strategy emulates retrieval augmented generation (RAG) \cite{RAG} to boost VLM reasoning. 
As illustrated in Figure {\color{brown} \ref{fig:VLM}}, GRE$^k_m$  first samples $k$ auxiliary neighbors, and then incorporates their multimodal attributes along with the target node’s via natural language instructions, akin to the \textit{Center-only} strategy.
We evaluate three retrieval settings: textual-only, visual-only, and both modalities. 
In detail, $m \in \left\{\{\mathcal{M}_\mathcal{T}\}, \{\mathcal{M}_\mathcal{V}\}, \{\mathcal{M}_\mathcal{T}, \mathcal{M}_\mathcal{V}\}\right\}$ denotes the retrieved modality.

\section{Experiment and Analysis} \label{Experiments}
In this section, we will first introduce the baselines and evaluation metrics adopted in experiments, followed by sufficient results and thorough analyses towards the two MAGRL paradigms \textit{GNN-as-Predictor} and \textit{VLM-as-Predictor}. Eventually, we provide insights on several up-and-coming MAG research directions.

\subsection{Experiment Setup}
\subsubsection{\textbf{Baseline}}
\textit{Firstly}, to investigate the impact of different attribute modalities on downstream GNNs, we adopt four widely used GNN models \cite{yan2023comprehensive}, i.e., GCN~\cite{GCN}, GraphSAGE~\cite{SAGE}, GAT~\cite{GAT}, and RevGAT~\cite{RevGAT}.
For multimodal embeddings, we leverage VLMs LLaMA3.2-11B Vision (LLaMAVL) and QWen2VL-7B (QWenVL) as the encoders.
To compare with the VLM embeddings, we concatenate the text embeddings of LLaMA3.1-8B and the visual embeddings of CLIP, as a kind of straightforward multimodal embedding.
For textual embeddings, we utilize RoBERTa~\cite{Roberta}, which is a lightweight model for text classification, along with typical LLMs such as LLaMA3.2-1B and LLaMA3.1-8B~\cite{llamav2}, and the textual branches of LLaMAVL and QWenVL.
For visual embeddings, we adopt different pre-trained vision models, including CLIP~\cite{CLIP}, ImageNet-Swinv2~\cite{Swinv2}, FCMAE-ConvNextV2~\cite{convnextv2}, and the visual branch of aforementioned VLMs. \textit{Secondly}, to study VLMs' zero-shot reasoning ability on MAG tasks, we adopt LLaMAVL and QWenVL as the backbone and evaluate with \textit{Center-only }and \textit{Graph retrieval modality enhancement} strategies. In this work, we focus on MAG node classification.
The generation details of different attribute embeddings are presented in Appendix \ref{Appendix-A}.



\subsubsection{\textbf{Evaluation metric}}
For the node classification task, we evaluate performance using \textit{Accuracy} and the \textit{F1-Macro} score. For the link prediction task, we use \textit{Hits@1}, \textit{Hits@3}, and \textit{Mean Reciprocal Rank} (MRR). We report the average result over 10-fold repetition for node classification and 5-fold repetition for link prediction, with different random seeds. 
Moreover, since VLMs may generate responses which is not exactly matched with the predefined candidate categories, we additionally define the \textit{Mismatch Rate} metric as follows, to quantify this phenomenon,
\begin{equation}
    \text{\textit{Mismatch Rate}}=\frac{1}{N}\sum_{n=1}^{N}\mathbb{I}(p_{\mathrm{model}}^{(n)}\notin P_{\mathrm{true}}^{(n)}),
\end{equation}
where \(N\) denotes the total node number, \(p_{\mathrm{model}}^{(n)}\) represents the  prediction output by the VLMs for node \(n\), and \(P_{\mathrm{true}}^{(n)}\) is the predefined category set.
One can refer to the Appendix \ref{app:spliting} for the dataset split.

\subsection{Performance Comparison of GNN}\label{sec:3.2}
Tables  {\color{brown}\ref{tab:NC of TE-VE-GNN}-\ref{tab:linkprediction}} present the experimental results of various GNN methods under different learning scenarios.
Complete experimental results are presented in the Appendix \ref{app:experiments}.
Our key findings are as follows:

\begin{table}[t]
  \centering
  \setlength{\abovecaptionskip}{0.2cm}
  \small
  \captionsetup{skip=5pt} 
  \caption{Basic statistics of MAGs in the MAGB.The variable \( L_{\rm text} \) represents the average number of words, while \( R_{\rm img} \) denotes the resolution of the image (in pixels).}
    \begin{tabular}{l|ccccc}
    \hline
    Dataset & Movies & Toys & Grocery & Reddit-S & Reddit-M \\
    \hline
    \#Nodes & 16,672 & 20,695 & 17,074 & 15,894 & 99,638 \\
    \#Edges & 218,390 & 126,886 & 171,340 & 566,160 & 1,167,188 \\
    \#Class & 20 & 18 & 20 & 20 & 50 \\
    $L_{\rm text}$ & 81.85 & 74.5 & 67.36 & 10.23 & 10.22 \\
    $R_{\rm img}$ & 388*476 & 467*442 & 402*457 & 2515*2399 & 2605*2662 \\
    \hline
    \end{tabular}%
  \label{tab:datasets}%
\vspace{-0.5cm}
\end{table}%


\ding{172} \textbf{Modality preferences differ by domain: social networks favor visual features, while e-commerce networks rely on textual attributes.}
As shown in Table {\color{brown}\ref{tab:NC of TE-VE-GNN}}, on the Reddit-M dataset, visual embeddings consistently outperform textual embeddings across all downstream GNNs and learning scenarios, with even relatively weaker vision models (e.g., ConvNeXT and SwinV2) surpassing textual embeddings derived from various large language models. This suggests that high-resolution visual content in social networks plays a more critical role than textual descriptions, which are typically shorter and less informative.
In contrast, on the Toys dataset, textual embeddings dominate across all settings, aligning with the characteristics of e-commerce networks, where product descriptions tend to be longer and contain more discriminative information, whereas product images are often of lower resolution and contribute less to classification. These findings reinforce the notion that modality importance varies significantly across different domains, influenced by data characteristics and content richness.


\begin{table*}[t]
  \centering
  \captionsetup{skip=5pt} 
  \caption{The node classification experimental results of TE-GNN and VE-GNN on the Movies, Toys, and Reddit-M datasets. Best results are highlighted with \colorbox[HTML]{52B788}{\textbf{bold}}, the second are marked with \colorbox[HTML]{95D5B2}{\underline{underline}}, and the third highest results are \colorbox[HTML]{D8F3DC}{marked}.}
    \begin{tabular}{clcccccccccc}
    \hline
    \multirow{2}[2]{*}{Settings} & \multicolumn{1}{c}{\multirow{2}[2]{*}{Feature}} & \multicolumn{5}{c}{Text} & \multicolumn{5}{c}{Visual}\\
      &   & \multicolumn{1}{l}{PLM} & \multicolumn{1}{l}{LLaMA1B} & \multicolumn{1}{l}{LLaMA8B} & \multicolumn{1}{l}{QWenVL} & \multicolumn{1}{l}{LLaMAVL} & \multicolumn{1}{l}{CLIP} & \multicolumn{1}{l}{ConvNeXT} & \multicolumn{1}{l}{Swin} & \multicolumn{1}{l}{QWenVL} & \multicolumn{1}{l}{LLaMAVL}\\
			\midrule
			\rowcolor{gray!8}\multicolumn{12}{c}{\textit{Dataset: Movies. Metric: Accuracy}}\\
			\midrule
    \multicolumn{1}{c}{\multirow{4}[2]{*}{Full}} & GCN & 52.44 & 53.18 & \third{53.90} & 53.43 & 53.79 & 52.85 & 51.30 & 51.82 & \first{\textbf{54.44}} & \second{\underline{54.26}}\\
      & SAGE & 53.81 & 55.20 & 55.73 & 54.96 & \third{55.76} & 54.08 & 51.27 & 51.54 & \first{\textbf{56.89}} & \second{\underline{56.14}} \\
      & GAT & 51.42 & 52.10 & \third{53.26} & 52.37 & \first{\textbf{53.37}} & 52.42 & 51.69 & 51.57 & \second{\underline{53.31}} & 51.36 \\
      & RevGAT & 54.04 & 55.06 & \third{56.12} & 55.02 & 55.95 & 54.43 & 50.80 & 51.32 & \first{\textbf{57.46}}& \second{\underline{56.76}} \\
    \hline
    \multicolumn{1}{c}{\multirow{4}[2]{*}{10-shot}} & GCN & 28.63 & \second{\underline{30.42}} & 23.95 & 26.00 & 23.98 & 28.97 & 25.81 & 25.86 & \third{29.13} & \first{\textbf{31.91}}\\
      & SAGE & 26.00 & 25.38 & 22.93 & 23.08 & 22.43 & \third{26.53} & 22.35 & 23.17 & \first{\textbf{28.30}} & \second{\underline{27.68}} \\
      & GAT & 26.87 & \second{\underline{28.87}} & \third{26.92} & 24.61 & 25.25 & 26.34 & 24.34 & 24.42 & 26.90 & \first{\textbf{29.60}} \\
      & RevGAT & 22.14 & \third{24.56} & 23.35 & 22.45 & 22.69 & 23.30 & 17.06 & 17.89 & \second{\underline{25.72}} & \first{\textbf{28.25}} \\
    \hline
    \multicolumn{1}{c}{\multirow{4}[2]{*}{ 3-shot}} & GCN & 17.10 & \second{\underline{20.27}} & 17.54 & 17.42 & \third{17.95} & 15.19 & 14.98 & 14.73 & 16.22 & \first{\textbf{21.02}} \\
      & SAGE & 19.57 & \first{\textbf{23.75}} & 17.92 & 17.26 & 18.04 & 16.99 & 14.81 & 13.98 & \second{\underline{19.82}} & \third{19.63} \\
      & GAT & 19.53 & \first{\textbf{23.43}} & \third{19.95} & 17.59 & 19.41 & 16.08 & 17.75 & 15.64 & 19.19 & \second{\underline{20.93}} \\
      & RevGAT & 14.33 & \second{\underline{17.98}} & 16.01 & 15.48 & 16.11 & 16.61 & 13.83 & 13.96 & \third{16.99} & \first{\textbf{18.81}} \\
			\midrule
			\rowcolor{gray!8}\multicolumn{12}{c}{\textit{Dataset: Toys. Metric: Accuracy}}\\
			\midrule
    \multicolumn{1}{c}{\multirow{4}[2]{*}{Full}} & GCN & 80.10 & 81.16 & \second{\underline{81.55}} & \third{81.43} & \first{\textbf{81.62}} & 79.31 & 78.83 & 78.73 & 80.45 & 80.18 \\
      & SAGE & 80.39 & 81.84 & \second{\underline{82.30}} & \third{82.27} & \first{\textbf{82.36}} & 79.50 & 79.03 & 78.73 & 80.88 & 80.30 \\
      & GAT & 79.88 & 80.85 & \first{\textbf{81.01}} & \third{81.00} & \second{\underline{81.01}} & 79.48 & 79.13 & 79.12 & 80.16 & 79.75 \\
      & RevGAT & 80.54 & 82.33 & \second{\underline{82.90}} & \third{82.88} & \first{\textbf{82.92}} & 79.63 & 78.03 & 77.91 & 81.09 & 79.84 \\
    \hline
    \multicolumn{1}{c}{\multirow{4}[2]{*}{ 10-shot}} & GCN & 57.45 & 60.85 & \third{62.96} & \first{\textbf{63.37}} & \second{\underline{62.99}} & 57.96 & 52.80 & 52.70 & 61.23 & 60.20 \\
      & SAGE & 55.19 & 59.23 & \first{\textbf{62.05}} & \second{\underline{61.89}} & \third{61.71} & 59.62 & 52.01 & 52.86 & 61.01 & 61.34 \\
      & GAT & 57.94 & 60.33 & \first{\textbf{61.79}} & \third{61.50} & \second{\underline{61.51}} & 58.02 & 52.24 & 52.22 & 60.62 & 60.66 \\
      & RevGAT & 56.17 & 59.74 & \second{\underline{62.15}} & \third{61.32} & \first{\textbf{62.43}} & 57.04 & 47.45 & 47.13 & 59.16 & 57.37 \\
    \hline
    \multicolumn{1}{c}{\multirow{4}[2]{*}{ 3-shot}} & GCN & 41.31 & 48.20 & 50.14 & \second{\underline{50.27}} & \third{50.22} & 43.59 & 39.60 & 37.73 & \first{\textbf{50.96}} & 49.48 \\
      & SAGE & 40.24 & 45.48 & 49.47 & \second{\underline{50.49}} & 48.62 & 46.12 & 36.54 & 36.89 & \third{50.08} & \first{\textbf{51.30}} \\
      & GAT & 42.77 & 45.14 & \second{\underline{47.96}} & \third{47.59} & 47.41 & 41.00 & 36.95 & 34.92 & 47.24 & \first{\textbf{50.76}} \\
      & RevGAT & 35.55 & 43.14 & \second{\underline{47.46}} & \first{\textbf{47.55}} & \third{46.99} & 39.47 & 31.93 & 29.24 & 43.24 & 44.18 \\
			\midrule
			\rowcolor{gray!8}\multicolumn{12}{c}{\textit{Dataset: Reddit-M. Metric: F1-Macro}}\\
			\midrule
    \multicolumn{1}{c}{\multirow{4}[2]{*}{Full}} & GCN & 64.92 & 67.43 & 68.29 & 68.04 & 68.32 & \first{\textbf{71.94}} & 70.45 & 70.73 & \third{71.81} & \second{\underline{71.84}} \\
      & SAGE & 69.13 & 72.77 & 74.83 & 74.00 & 74.87 & \third{79.61} & 77.08 & 77.84 & \second{\underline{79.66}} & \first{\textbf{79.66}} \\
      & GAT & 65.49 & 67.54 & 68.14 & 67.41 & 68.17 & \first{\textbf{71.16}} & 69.39 & 69.92 & \third{70.34} & \second{\underline{70.76}} \\
      & RevGAT & 70.38 & 73.79 & 75.55 & 74.55 & 75.55 & \first{\textbf{80.29}} & 77.61 & 78.31 & \third{79.90} & \second{\underline{80.15}} \\
    \hline
    \multicolumn{1}{c}{\multirow{4}[2]{*}{ 10-shot}} & GCN & 31.74 & 41.95 & 46.78 & 44.23 & 46.38 & \second{\underline{67.61}} & 64.91 & 65.12 & \third{67.20} & \first{\textbf{67.65}} \\
      & SAGE & 30.76 & 39.91 & 44.57 & 41.77 & 44.07 & \second{\underline{65.40}} & 61.52 & 61.46 & \third{64.70} & \first{\textbf{66.21}} \\
      & GAT & 32.79 & 37.99 & 40.86 & 35.94 & 39.38 & \first{\textbf{66.85}} & 60.60 & 59.37 & \third{64.09} & \second{\underline{65.64}} \\
      & RevGAT & 32.13 & 40.20 & 43.04 & 40.59 & 43.19 & \third{63.28} & 60.86 & 59.96 & \second{\underline{64.11}} & \first{\textbf{66.29}} \\
    \hline
    \multicolumn{1}{c}{\multirow{4}[2]{*}{ 3-shot}} & GCN & 20.19 & 27.62 & 31.72 & 29.36 & 31.63 & \first{\textbf{59.77}} & 58.21 & 58.39 & \third{58.59} & \second{\underline{59.54}} \\
      & SAGE & 17.22 & 23.81 & 27.42 & 25.83 & 27.46 & \second{\underline{56.29}} & 54.64 & 54.22 & \third{55.26} & \first{\textbf{56.62}} \\
      & GAT & 17.82 & 20.89 & 22.86 & 20.61 & 22.96 & \first{\textbf{57.71}} & \third{52.05} & 49.18 & 51.82 & \second{\underline{55.67}} \\
      & RevGAT & 16.42 & 24.75 & 26.44 & 24.17 & 26.22 & \third{53.26} & \second{\underline{53.64}} & 51.30 & 52.36 & \first{\textbf{56.37}} \\
    \hline
    \end{tabular}%
  \label{tab:NC of TE-VE-GNN}%
\end{table*}%

\begin{table*}[t]
  \centering
   \captionsetup{skip=5pt} 
  \caption{Comparison of ME-GNN with TE-GNN and VE-GNN on the node classification task under different settings.}
  \begin{adjustbox}{width=0.94\textwidth}
    \begin{tabular}{clcccc|cccc|cccc}
    \hline
    \multirow{2}[2]{*}{Datasets} & \multicolumn{1}{c}{\multirow{2}[2]{*}{Feature}} & \multicolumn{4}{c|}{Full} & \multicolumn{4}{c|}{10-Shot} & \multicolumn{4}{c}{3-Shot} \\
      &   & GCN & SAGE & GAT & RevGAT & GCN & SAGE & GAT & RevGAT & GCN & SAGE & GAT & RevGAT \\
    \hline
    \multirow{5}[2]{*}{Movies} & Text & 53.90 & 55.76 & 53.37 & 56.12 & \underline{30.42} & 26.00 & \underline{28.87} & 24.56 & \underline{20.27} & \textbf{23.75} & \textbf{23.43} & \underline{17.98} \\
      & Visual & 54.44 & 56.89 & 53.31 & 57.46 & \textbf{31.91} & \underline{28.30} & \textbf{29.60} & \textbf{28.25} & \textbf{21.02} & 19.82 & 20.93 & \textbf{18.81} \\
      & LLaMA+CLIP & 53.98 & 56.08 & 53.52 & 56.54 & 24.02 & 21.98 & 24.72 & 23.35 & 17.80 & 17.68 & 17.54 & 17.02 \\
      & QwenVL & \textbf{55.19} & \textbf{58.01} & \underline{53.93} & \underline{58.24} & 28.42 & \textbf{28.99} & 28.35 & \underline{27.43} & 16.41 & \underline{19.90} & \underline{22.35} & 17.34 \\
      & LLaMAVL & \underline{54.92} & \underline{57.39} & \textbf{53.94} & \textbf{58.71} & 26.11 & 25.44 & 26.12 & 25.30 & 16.76 & 18.34 & 18.72 & 17.05 \\
    \hline
    \multirow{5}[2]{*}{Toys} & Text & \textbf{81.62} & \textbf{82.36} & 81.01 & \underline{82.92} & \textbf{63.37} & 62.05 & \underline{61.79} & 62.43 & 50.27 & 50.49 & 47.96 & 47.55 \\
      & Visual & 80.45 & 80.88 & 80.16 & 81.09 & 61.23 & 61.34 & 60.66 & 59.16 & 50.96 & \underline{51.30} & \textbf{50.76} & 44.18 \\
      & LLaMA+CLIP & \underline{81.59} & 82.35 & \underline{81.01} & \textbf{83.03} & \underline{62.93} & \underline{62.83} & 61.66 & \underline{62.67} & 50.14 & 49.65 & 48.36 & \underline{47.69} \\
      & QwenVL & 80.95 & 81.45 & 80.64 & 81.99 & 62.78 & 62.81 & 61.77 & 61.34 & \textbf{52.83} & \textbf{52.13} & 48.02 & 47.64 \\
      & LLaMAVL & 81.57 & \underline{82.35} & \textbf{81.13} & 82.88 & 62.70 & \textbf{63.34} & \textbf{61.88} & \textbf{63.82} & \underline{51.95} & 51.20 & \underline{49.43} & \textbf{49.74} \\
    \hline
    \multirow{5}[2]{*}{Reddit-M} & Text & 68.32 & 74.87 & 68.17 & 75.55 & 46.78 & 44.57 & 40.86 & 43.19 & 31.72 & 27.46 & 22.96 & 26.44 \\
      & Visual & 71.94 & 79.66 & 71.16 & 80.29 & \underline{67.65} & \underline{66.21} & \underline{66.85} & \underline{66.29} & \underline{59.77} & \underline{56.62} & \textbf{57.71} & \textbf{56.37} \\
      & LLaMA+CLIP & 71.42 & 80.84 & \underline{71.56} & 80.66 & 48.39 & 45.70 & 40.38 & 45.06 & 32.61 & 28.10 & 23.92 & 27.03 \\
      & QwenVL & \underline{72.96} & \underline{81.23} & 71.48 & \underline{81.32} & 67.29 & 65.59 & 63.85 & 64.07 & 58.95 & 55.74 & 51.36 & 52.54 \\
      & LLaMAVL & \textbf{75.64} & \textbf{85.61} & \textbf{74.61} & \textbf{86.06} & \textbf{71.09} & \textbf{70.16} & \textbf{67.37} & \textbf{68.95} & \textbf{61.97} & \textbf{57.36} & \underline{53.40} & \underline{54.31} \\
    \hline
    \end{tabular}%
  \label{tab:NC of MEGNN}%
  \end{adjustbox}
  \vspace{-5pt}
\end{table*}%

\begin{table*}[htbp]
  \centering
  \captionsetup{skip=5pt} 
  \caption{Comparison of ME-GNN with TE-GNN and VE-GNN on the link prediction task}
  \begin{adjustbox}{width=0.94\textwidth}
    \begin{tabular}{lccc|ccc|ccc|ccc|ccc}
    \hline
    \multicolumn{1}{c}{\multirow{2}[2]{*}{Feature}} & \multicolumn{3}{c}{Movies} & \multicolumn{3}{c}{Toys} & \multicolumn{3}{c}{Grocery} & \multicolumn{3}{c}{Reddit-S} & \multicolumn{3}{c}{Reddit-M} \\
      & Hits1 & Hits3 & \multicolumn{1}{c}{MRR} & Hits1 & Hits3 & \multicolumn{1}{c}{MRR} & Hits1 & Hits3 & \multicolumn{1}{c}{MRR} & Hits1 & Hits3 & \multicolumn{1}{c}{MRR} & Hits1 & Hits3 & MRR \\
			\midrule
			\rowcolor{gray!8}\multicolumn{16}{c}{\textit{Backbone model: GCN}}\\
			\midrule
    RoBERTa & 12.78 & 35.25 & 30.35 & 16.85 & 41.90 & 34.79 & 12.83 & 36.07 & 30.09 & 2.88 & 15.21 & 16.80 & \second{\underline{59.18}} & 86.87 & \first{\textbf{76.82}} \\
      LLaMA1B & 13.30 & 36.75 & 31.19 & 18.09 & 44.50 & 36.90 & 14.49 & 38.22 & 32.33 & 2.70 & 14.54 & 16.41 & 56.58 & 86.32 & 75.21 \\
      LLaMA8B & 13.41 & 37.00 & 31.31 & 18.39 & 45.34 & 37.26 & 14.63 & 39.06 & \first{\textbf{32.68}} & 2.74 & 14.82 & 16.56 & 58.88 & 86.68 & 76.61 \\
      QWenVL-T & \first{\textbf{13.45}} & 36.82 & 31.28 & 18.19 & 45.51 & 37.17 & 14.55 & \second{\underline{39.32}} & 32.45 & 2.80 & 14.95 & 16.58 & 58.96 & 86.74 & 76.62 \\
      LLaMAVL-T & 13.37 & \second{\underline{37.25}} & 31.34 & 18.36 & 45.15 & 37.47 & 14.48 & 39.00 & 32.64 & 2.63 & 14.82 & 16.55 & 58.87 & 86.70 & 76.62 \\
      \hline
      CLIP & 13.35 & 36.65 & 31.40 & 18.07 & 43.60 & 36.17 & 13.66 & 36.52 & 30.69 & \first{\textbf{3.56}} & \first{\textbf{17.74}} & \first{\textbf{18.09}} &  \first{\textbf{59.27}} & 86.46 & 76.63 \\
      ConvNext & 13.35 & 36.75 & 31.05 & 17.94 & 44.94 & 36.53 & 14.12 & 37.95 & 31.83 & \second{\underline{2.94}} & \second{\underline{15.63}} & \second{\underline{16.96}} & 58.72 & 86.21 & 76.32 \\
      Swin  & 13.30 & 36.99 & 31.31 & 18.34 & \second{\underline{46.05}} & 37.11 & 14.05 & 38.83 & 32.03 & 2.84 & 15.49 & 16.89 & 58.73 & 86.33 & 76.36 \\
      QWenVL-V & 13.41 & 36.95 & 31.28 & 18.18 & 45.41 & 37.38 & 14.50 & 38.15 & 32.09 & 2.45 & 14.11 & 16.10 & 56.73 & 85.69 & 75.00 \\
      LLaMAVL-V & 12.63 & 34.83 & 29.93 & 15.93 & 38.92 & 33.14 & 13.00 & 34.32 & 29.53 & 1.78 & 10.07 & 13.52 & 44.05 & 77.69 & 65.12 \\
      \hline
      LLaMA+CLIP & \second{\underline{13.44}} & 36.97 & 31.45 & \first{\textbf{18.58}} & 45.62 & 37.51 & \first{\textbf{14.72}} & 39.10 & \second{\underline{32.67}} & 2.76 & 14.79 & 16.58 & 59.10 & \first{\textbf{86.91}} & \second{\underline{76.75}} \\
      QWenVL & 13.30 & \first{\textbf{37.43}} & \first{\textbf{31.69}} & 18.38 & \first{\textbf{46.31}} & \first{\textbf{37.87}} & 14.55 & 38.63 & 32.41 & 2.63 & 14.65 & 16.44 & 57.86 & 86.53 & 76.10 \\
      LLaMAVL & 13.42 & 36.98 & \second{\underline{31.66}} & \second{\underline{18.51}} & 45.99 & \second{\underline{37.65}} & \second{\underline{14.68}} & \first{\textbf{39.35}} & 32.61 & 2.74 & 14.77 & 16.45 & 58.87 & \second{\underline{86.88}} & 76.64 \\
      	\midrule
		\rowcolor{gray!8}\multicolumn{16}{c}{\textit{Backbone model: MLP}}\\
		\midrule
     RoBERTa & 3.57 & 10.59 & 10.11 & 7.12 & 17.49 & 16.06 & 4.05 & 11.57 & 10.41 & 1.16 & 4.55 & 5.87 & 4.32 & 9.00 & 9.13 \\
    LLaMA1B & 6.32 & 19.11 & 17.43 & 10.43 & 28.48 & 24.78 & 6.92 & 20.80 & 18.52 & 1.78 & 6.14 & 8.42 & 6.72 & 15.15 & 14.54 \\
    LLaMA8B & 7.37 & 22.97 & 20.30 & \second{\underline{11.83}} & 31.55 & 27.44 & 8.05 & \second{\underline{24.05}} & 21.41 & 2.26 & 7.80 & 10.09 & 8.76 & 19.33 & 18.24 \\
    QWenVL-T & 6.59 & 20.36 & 18.35 & 10.49 & 30.06 & 25.86 & 6.99 & 21.61 & 18.97 & 1.88 & 6.48 & 8.74 & 7.58 & 16.95 & 16.31 \\
    LLaMAVL-T & 7.55 & 22.80 & 20.50 & 11.56 & 32.06 & 27.46 & 7.43 & 23.46 & 21.28 & 2.21 & 7.56 & 9.96 & 8.45 & 19.20 & 18.04 \\
    \hline
    CLIP & 6.55 & 18.06 & 16.45 & 7.33 & 20.53 & 17.64 & 4.37 & 12.85 & 11.50 & 1.90 & 6.65 & 8.28 & 10.94 & 22.01 & 20.63 \\
    ConvNext & 4.03 & 12.95 & 12.80 & 5.96 & 17.88 & 16.31 & 4.04 & 13.64 & 13.07 & \first{\textbf{3.42}} & \first{\textbf{10.84}} & \first{\textbf{12.72}} & 12.02 & 25.41 & 24.16 \\
    Swin  & 3.58 & 11.71 & 12.08 & 5.75 & 17.32 & 16.18 & 3.59 & 12.40 & 12.22 & \second{\underline{2.87}} & \second{\underline{9.48}} & 11.38 & 9.04 & 19.98 & 19.42 \\
    QWenVL-V & 6.61 & 20.71 & 18.59 & 9.29 & 24.92 & 22.29 & 4.74 & 14.57 & 13.58 & 1.98 & 6.73 & 9.01 & 10.35 & 22.60 & 21.21 \\
    LLaMAVL-V & 6.31 & 18.03 & 16.39 & 7.94 & 21.51 & 19.13 & 4.63 & 13.67 & 12.25 & 1.49 & 5.66 & 7.55 & 8.50 & 18.39 & 18.09 \\
    \hline
    LLaMA+CLIP & \second{\underline{7.56}} & \second{\underline{23.19}} & \second{\underline{20.77}} & 11.80 & \second{\underline{32.50}} & \second{\underline{27.99}} & \first{\textbf{8.09}} & 23.80 & \second{\underline{21.66}} & 2.37 & 8.27 & 10.52 & \second{\underline{12.44}} & \second{\underline{26.05}} & \second{\underline{24.18}} \\
    QWenVL & 7.51 & 23.07 & 20.41 & 10.64 & 28.86 & 25.11 & 6.16 & 18.51 & 16.96 & 2.00 & 7.06 & 9.25 & 10.76 & 23.43 & 21.78 \\
    LLaMAVL & \first{\textbf{8.22}} & \first{\textbf{24.98}} & \first{\textbf{22.29}} & \first{\textbf{12.82}} & \first{\textbf{34.48}} & \first{\textbf{29.47}} & 8.07 & \first{\textbf{25.12}} & \first{\textbf{21.81}} & 2.67 & 8.99 & \second{\underline{11.52}} & \first{\textbf{13.05}} & \first{\textbf{27.94}} & \first{\textbf{25.69}} \\
    \hline
    \end{tabular}%
  \label{tab:linkprediction}%
  \end{adjustbox}
\end{table*}%

\ding{173} \textbf{When training samples are abundant, multimodal embeddings substantially improve the performance of downstream GNNs. Conversely, in scenarios with limited training samples, the presence of modality bias may impede their effectiveness.} Table {\color{brown}\ref{tab:NC of MEGNN}} shows that in the supervised learning setting, multimodal embeddings outperform single-modal embeddings in 10 out of 12 cases across four GNN models and three datasets. For instance, on the Movies dataset, multimodal embeddings improve upon the best single-modal embeddings by 0.75\%, 1.12\%, 0.57\%, and 1.25\% across four GNN architectures. Similarly, in the Reddit-M dataset, improvements range from 3.7\% to 5.95\%.
However, in situations with limited training samples, modality bias within the multimodal embeddings may prevent GNNs from fully leveraging the diverse attributes, thereby diminishing their overall effectiveness. 
For example, in the Movies dataset, the best node features for the four GNNs under the 3-shot setting come from single-modal embeddings. The best multimodal embeddings are, on average, 2.41\% lower than the best single-modal embeddings.


\begin{table*}[t]
	\centering
    \captionsetup{skip=5pt} 
        \caption{
		Zero-shot node classification results on various VLMs. The best results are highlighted in \textbf{bold}, while the second-best results are \underline{underlined}. For the \underline{Mis}match rate, lower values indicate better performance.
	}
    \small 
	\begin{adjustbox}{width=0.95\textwidth}
		\begin{tabular}{lccccccccccccccc}
			\toprule
    \multicolumn{1}{c}{\multirow{2}{*}{Methods}} & \multicolumn{3}{c}{Movies} & \multicolumn{3}{c}{Toys} & \multicolumn{3}{c}{Grocery} & \multicolumn{3}{c}{Reddit-S} & \multicolumn{3}{c}{Reddit-M} 
    			\\\cmidrule{2-16}
			   & Acc & F1 & Mis & Acc & F1 & Mis & Acc & F1 & Mis & Acc & F1 & Mis & Acc & F1 & Mis \\
			\midrule
			\rowcolor{gray!8}\multicolumn{16}{c}{\textit{LLaMA-3.2 11B Vision Instruct}}\\
			\midrule
			{Center-only} & \underline{19.00} & 13.96 & \textbf{0.67} & \textbf{50.00} & \underline{40.56}  & \underline{6.33}  & \textbf{54.00} & \textbf{48.32} & \textbf{0.67} & \underline{74.00} & \underline{62.56} & \textbf{0.33}  & 54.67 & 46.55 & \textbf{3.67} \\
			{GRE-T$_{k=1}$} & 16.33 & 10.56 & 3.67 & \underline{48.00} & \textbf{41.79} & \textbf{1.00} & 48.33 & 39.69 & \underline{1.00} & 65.67 & 55.95 & 1.67 & 58.00 & 50.47 & 5.33\\
			{GRE-V$_{k=1}$} & 17.00 & 14.01 & \textbf{0.67} & 41.67 & 35.53 & 10.33 & 50.33 & 43.06 & \underline{1.00} & 68.00 & 57.63 & \underline{1.00} & 54.67 & 44.86 & \underline{4.00} \\
			{GRE-M$_{k=1}$} & 12.67 & 12.39 & \underline{2.67} & 38.00 & 33.43 & 11.00 & 47.00 & 37.62 & \textbf{0.67} & 58.33 & 48.17 & 1.33 & 49.67 & 41.72 & 6.00 \\
			{GRE-T$_{k=3}$} & \textbf{19.67} & \textbf{20.27} & \underline{2.67} & 45.67 & 39.36 & 10.67 & 51.00 & 43.29 & \textbf{0.67} & 72.00 & 62.29 & 2.33 &  \underline{60.33} & \textbf{51.65} & 4.67 \\
			{GRE-T$_{k=5}$} & 18.33 &  \underline{15.56} & 3.00 & 42.67 & 37.67 & 11.67 &  \underline{52.00} & 45.73 & 2.33 & \textbf{74.67} & \textbf{62.70} & \textbf{0.33} & \textbf{61.67} &  \underline{51.45} & 5.33 \\
			\midrule
			\rowcolor{gray!8}\multicolumn{16}{c}{\textit{Qwen2-VL 7B Instruct}}\\
			\midrule
			{Center-only} &   5.00 & 6.11 & 41.33 & 42.67 & 37.48 & 18.33 &  \underline{30.33} &  \underline{29.99} & \underline{46.33} & 59.67 & 57.92 & 28.67 & 53.33 & 48.55 & 25.67 \\
			{GRE-T$_{k=1}$} &     5.33 & 4.97 & 42.67 &  \underline{51.00} &  \underline{43.93} & 17.33 & 30.00 & 25.66 & 49.00 & 58.33 & 58.10 & 28.67 & 56.00 & 50.60 & 22.33 \\
			{GRE-V$_{k=1}$} &   \textbf{9.00} & \textbf{11.18} & 54.00 & 46.00 & 39.37 & \textbf{15.67} & 27.33 & \textbf{31.86} & 54.33 & 58.67 & 58.15 & 28.67 & 53.67 & 48.40 & 24.00 \\
			{GRE-M$_{k=1}$} &   4.67 & 7.15 & 63.00 & 43.00 & 38.20 & 26.00 & 24.33 & 25.67 & 53.33 & 53.67 & 55.76 & 35.33 & 46.00 & 44.05 & 29.33 \\
			{GRE-T$_{k=3}$} &   \underline{7.33} &  \underline{8.25} & \textbf{37.33} & 50.33 & 43.57 & \underline{16.33} & 29.33 & 26.55 & 46.33 & \textbf{63.67} & \textbf{61.70} & \underline{23.67} & \textbf{57.67} & \textbf{51.35} & \textbf{20.33} \\
			{GRE-T$_{k=5}$} &  \underline{7.33} & 6.12 & \underline{38.00} & \textbf{51.33} & \textbf{43.96} & \underline{16.33} & \textbf{30.67} & 27.81 & \textbf{43.67} &  \underline{63.33} &  \underline{61.02} & \textbf{23.00} &  \underline{57.33} &  \underline{51.17} & \underline{21.00} \\
			\bottomrule
		\end{tabular}
	\end{adjustbox}
	\label{exp:lavr-exp}
\end{table*}

\begin{figure*}[t]
  \centering
  \setlength{\abovecaptionskip}{0pt} 
  \includegraphics[width=0.98\textwidth]{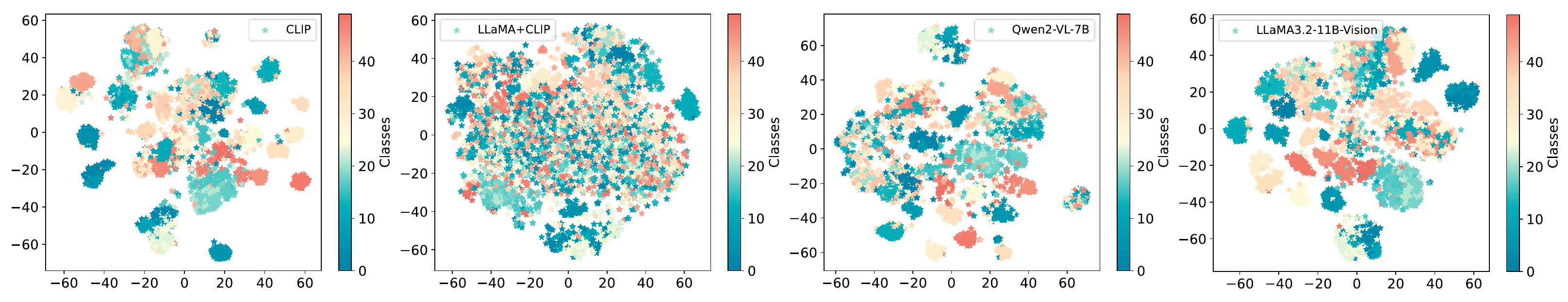}
  \caption{TSNE visualization of different embeddings on Reddit-M dataset.}
    \vspace{-10pt} 
  \label{fig:vlm-tsne}
\end{figure*}

\ding{174} \textbf{Large Vision Language Models (VLMs) are highly effective in generating multimodal embeddings, significantly improving performance and alleviating modality imbalance.} 
In Table {\color{brown}\ref{tab:NC of MEGNN}}, the multimodal embeddings generated by VLMs achieve the best performance in 31 out of 36 experimental scenarios. This indicates that generative VLMs serve as effective multimodal attribute encoders. Furthermore, we find that there is a significant disparity in the contribution of text and visual attributes on the Reddit-M dataset. In the 10-shot and 3-shot node classification experiments, the best visual embeddings outperform the best textual embeddings by 22.9\% and 30.47\%, respectively. When using the CONCAT method to fuse textual and visual features, the resulting multimodal embeddings completely lose the excellent clustering information of the visual representation due to significant differences in feature dimensions (with \textit{LLaMA-3.1-8B}'s textual features having a dimension of 4096, while CLIP's visual features have a dimension of 896, as shown in Figure {\color{brown}\ref{fig:vlm-tsne}
}). In contrast, the multimodal embeddings generated by VLMs (as shown in the right half of Figure {\color{brown}\ref{fig:vlm-tsne}
}) still retain good clustering information. Notably, based on the results in Table {\color{brown}\ref{tab:NC of TE-VE-GNN}}, the multimodal embeddings generated by \textit{LLaMA3.2-11B Vision} achieve an average improvement of 2.95\% over visual embeddings, despite the lower quality of textual embeddings. In the 3-shot scenario, multimodal embeddings improve the performance of GCN and GraphSAGE by 2.43\% and 0.74\%, respectively, while decreasing the performance of GAT and RevGAT by 2.27\% and 2.06\%, respectively. This highlights the need for further optimization of the multimodal embedding extraction process to better adapt to different types of GNNs.

\subsection{Evaluating Link Prediction Performance}
We evaluate the link prediction performance of attribute-enhanced models on MAGs by comparing GNN-based methods with MLP-based baselines to assess their effectiveness in leveraging structural and multimodal information.
Due to space constraints, Table {\color{brown}\ref{tab:linkprediction}} reports the results for GCN, which serves as a representative GNN for link prediction.
Based on these results, we summarize the following key findings. 
\textbf{\textit{(i)}} \textbf{GNN-based methods consistently outperform MLP-based methods} in link prediction tasks across five datasets. For example, in the Reddit-M dataset, the attribute-enhanced GCN achieved improvements of 46.22\% in Hits@1, 58.97\% in Hits@3, and 51.13\% in MRR compared to the MLP. This highlights the advantage of GNNs in leveraging graph structural information.
\textbf{\textit{(ii)}}\textbf{ Multimodal embeddings significantly enhance the performance of MLP-based methods,} suggesting that diverse data sources provide complementary information that improves model effectiveness. This indicates that even without explicit graph structures, multimodal attributes can help bridge the performance gap between MLP and GNN models.
\textbf{\textit{(iii)}}\textbf{ The influence of attribute features on GNN performance is less significant than in node classification}, likely due to the message-passing mechanism, which tends to smooth and aggregate node features, thereby reducing the impact of heterogeneous attributes. Additionally, while ConvNext demonstrated moderate performance in node classification, it performed comparably to CLIP in link prediction tasks, suggesting that the effectiveness of node representations is highly task-dependent.

\subsection{Zero-shot MAG Learning}
Furthermore, we investigate the zero-shot learning capacity of large vision language models (LVLMs) on multimodal attributed graphs (MAGs). Specifically, the evaluated LVLMs include \textit{LLaMA-3.2 11B Vision Instruct} and \textit{Qwen2-VL 7B Instruct}. According to the result presented in Table {\color{brown}\ref{exp:lavr-exp}}, one can draw the following conclusions.

\ding{172} \textbf{VLMs are competitive zero-shot MAG learners.}
One may notice that the zero-shot node classification performance in Table {\color{brown}\ref{exp:lavr-exp}} is comparable to the 3-shot performance shown in Table {\color{brown}\ref{tab:NC of MEGNN}}, especially in the Reddit-M, where the degradation is merely 0.48\%.

\ding{173} \textbf{Graph Retrieval Enhancement (GRE) is generally effective in facilitating VLMs on MAG learning.}
By introducing neighbor attributes, GRE can improve MAG node classification performance in most scenarios. For the \textit{Qwen2-VL-7B} model, the average gain of GRE reaches 4.27\% and 4.00\%.

\ding{174} \textbf{Multimodal attributes may disturb GRE in zero-shot situations.} 
Compared to relying solely on unimodal neighbor information, using both text and image information from neighbors during graph retrieval enhancement seems to decrease node classification performance.
For \textit{{LLaMA3.2-11B-Vision}}, the performance gap is more than 9.20\% and 7.72\%. This indicates that the VLM may struggle to effectively differentiate the importance of central node information from that of neighbor node information. 

\begin{figure}[t]
  \centering
  \setlength{\abovecaptionskip}{0pt} 
  \includegraphics[width=0.45\textwidth]{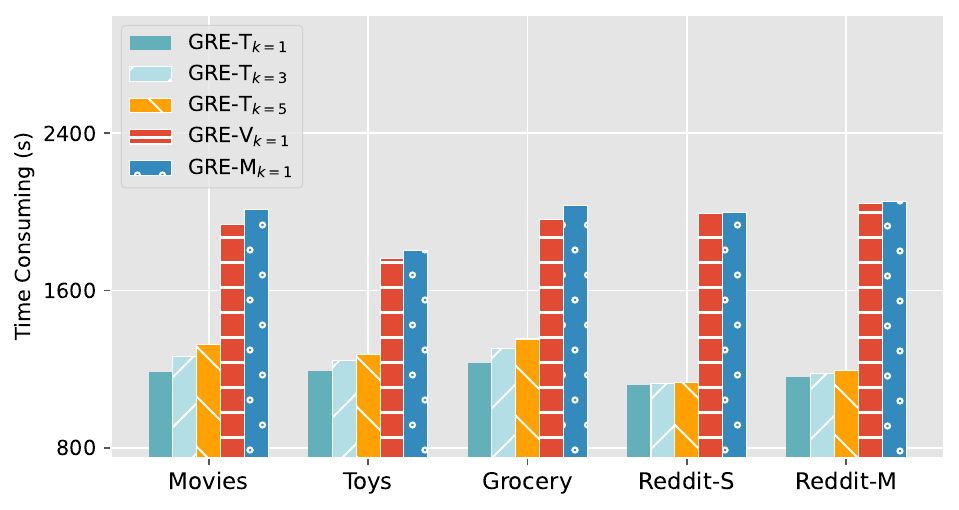}
  \caption{Time Consuming of LLaMA-3.2 11B Vision Model.}
    \vspace{-10pt} 
  \label{fig:time}
\end{figure}

\ding{175} \textbf{Increasing the number of neighbors brings nuanced benefits.} 
When enlarging the neighbor number (with $k$ set as 1, 3, and 5), the task performance consistently improves.
However, the marginal effect is also evident, as the performance of $k=5$ is not necessarily better than that of smaller $k$ values. 
This improvement in performance comes at a cost, as shown in Figure {\color{brown}\ref{fig:time}}, where the total reasoning time of the VLM gradually increases with the number of retrieved neighbors. Notably, retrieving images significantly increases the reasoning time compared to retrieving text.

\ding{176} \textbf{The uncontrollable output issue of VLMs may significantly impact their performance.}
Observing Table {\color{brown}\ref{exp:lavr-exp}}, it is evident that \textit{LLaMA3.2-VL}'s outputs are more controllable than \textit{Qwen2-VL}'s. For example, under the \textit{Center-only} strategy on the Movies dataset, \textit{LLaMA3.2-VL} reduces the Mis metric by 40.66\% compared to \textit{Qwen2-VL}.
In the Appendix \ref{app:uncontroll}, we will further discuss several situations in which \textit{Qwen2-VL}'s output is uncontrollable.

\subsection{Future Direction} \label{subsec:future_direction} Based on our analyses, we highlight the following key directions to enhance representation learning on MAGs:
\textbf{\textit{ (i) }Optimizing VLM Strategies for Multimodal Embeddings.}
Future research should focus on improving Vision-Language Models (VLMs) for extracting multi-modal representations, exploring alignment training and prompt-based techniques to enhance quality and reduce discrepancies.
\textbf{\textit{ (ii) }Advancing Multimodal GNNs.}
Developing Multi-modal Graph Neural Networks (M-GNNs) is crucial for effectively learning from graph data with rich attributes. Enhancing GNNs to integrate diverse multi-modal information will improve performance in tasks like node classification and link prediction.
\textbf{\textit{ (iii) }Exploring New Learning Paradigms.}
Investigating new paradigms for integrating GNNs and VLMs can enhance learning on MAGs. Collaborative strategies that leverage their strengths will improve multi-modal understanding.
\textbf{\textit{ (iv) }Enriching MAG Datasets.}
To improve multi-modal representation learning, future research should expand MAG datasets by incorporating diverse domains, adding new modalities (e.g., audio, and video), and developing dynamic datasets to capture evolving relationships.

\section{Related Work}
Many well-established graph benchmarks already exist in the graph community, especially for the Text-attributed graphs~\cite{OGB,Large,Amazon-Photo,GRL}. 
OGB \cite{OGB} provides large-scale and diverse datasets, laying the foundation for subsequent developments with datasets such as OGBN-arxiv, which incorporate textual attributes.
CS-TAG \cite{yan2023comprehensive} further expands the availability of TAG datasets and explores the impact of attribute modeling on TAG representation learning.
Building upon CS-TAG \cite{yan2023comprehensive}, TEG \cite{TEG} introduces graph datasets where both nodes and edges are enriched with textual descriptions, facilitating research and applications in Textual-Edge Graphs (TEGs).
DyTAG \cite{DTGB}, on the other hand, acknowledges the dynamic nature of real-world text-attributed graphs, where both graph structures and textual descriptions evolve over time. To advance research in this field, it provides benchmark datasets that accurately capture dynamic graph structures and rich textual attributes.
However, despite the progress made by these benchmarks, significant challenges remain in studying representation learning on MAGs.
Firstly, most existing datasets primarily focus on unimodal attributes and often lack raw, diverse attribute information~\cite{yan2023comprehensive}, making it difficult to explore the integration of multimodal attribute knowledge with structural information.
Secondly, the available multimodal attributed graph datasets often lack a standardized format, which hinders reproducibility and large-scale benchmarking efforts.
Thirdly, these benchmarks have not designed detailed learning paradigms to comprehensively compare the importance of multimodal attribute knowledge with graph structure for representation learning on MAGs.
Therefore, it is necessary to construct a comprehensive dataset and benchmark for MAGs.

\section{Conclusion}
We establish a comprehensive benchmark, \name, specifically designed to advance the exploration of Multimodal Attributed Graph (MAG) representation learning. This benchmark includes five standardized multimodal attributed graph datasets, enabling collaboration among the Graph, Natural Language Processing (NLP), and Computer Vision (CV) communities to collectively investigate the data.
Our benchmark facilitates a more thorough evaluation of various learning methods, highlighting their effectiveness and limitations. Furthermore, we will continue to identify and develop additional research-worthy MAGs to foster the sustained and healthy advancement of the field.


\bibliographystyle{ACM-Reference-Format}
\balance
\bibliography{main}

\clearpage



\appendix

\section{Model Details}\label{Appendix-A}
\subsection{Backbone Graph Neural Networks}
\label{app:Baselines} 
First, we describe the backbone graph neural networks used in the main experiments.

\textbf{GCN~\cite{GCN}.}  Graph Convolutional Network (GCN) is a classical model that works by performing a linear approximation to spectral graph convolutions. It uses an efficient layer-wise propagation rule that is based on a first-order approximation of spectral convolutions on graphs.

\textbf{GraphSAGE~\cite{SAGE}} is a GNN model that focuses on inductive node classification but can also be applied to transductive settings. It leverages node features to generate node embeddings for previously unseen data efficiently. Instead of training individual embeddings for each node, it learns a function that generates embeddings by sampling and aggregating features from a node's local neighborhood.

\textbf{GAT~\cite{GAT}.}  Graph Attention Network (GAT) introduces the attention mechanism to capture the importance of neighboring nodes when aggregating information from the graph. By stacking masked self-attentional layers in which nodes can attend over their neighborhoods’ features, a GAT specifies different weights to different nodes in a neighborhood without requiring any costly matrix operation or depending on knowing the graph structure upfront.

\textbf{RevGAT~\cite{RevGAT}}  generalizes reversible residual connections to grouped reversible residual connections. It extends the idea of reversible connections, which allow information to flow both forward and backward to vanilla GAT architecture. By doing so, RevGAT is able to improve the expressiveness and efficiency of GNNs.

\subsection{Textual Attribute Enconder}
For textual attribute modality, we use the pre-trained language models and generative large language models to get the textual embeddings of the nodes. 

\textbf{RoBERTa-Base} is a transformer-based language model developed by Facebook AI in 2019. It serves as an optimized version of the BERT \cite{bert} model, featuring 125 million parameters. RoBERTa-Base was trained on a large corpus of English data in a self-supervised manner, meaning it was pretrained on raw texts without human labeling. The model improves upon BERT by training longer with larger batches over more data, removing the next-sentence prediction objective, and utilizing dynamic masking patterns. These enhancements enable RoBERTa-Base to produce contextualized word representations, making it effective for various natural language processing tasks such as text classification, sentiment analysis, and question-answering.

\textbf{Llama3.2-1B-Instruct~\cite{llamav2}.} In September 2024, Meta introduced LLaMA-3.2-1B-Instruct, a model within the LLaMA 3.2 series of multilingual large language models. This model contains 1 billion parameters and has undergone instruction tuning to improve its performance in multilingual dialogue applications. It supports languages including English, German, French, Italian, Portuguese, Hindi, Spanish, and Thai. Optimized for tasks involving natural language understanding and generation, such as retrieval and summarization, LLaMA-3.2-1B-Instruct features a context window of 128,000 tokens, enabling it to manage extensive context for tasks like long-form text summarization.

\textbf{Llama3.1-8B-Instruct~\cite{llamav2}.} Released by Meta in July 2024, LLaMA-3.1-8B-Instruct is part of the LLaMA 3.1 series of multilingual large language models. This model comprises 8 billion parameters and has been instruction-tuned to enhance performance in multilingual dialogue applications. It supports languages such as English, German, French, Italian, Portuguese, Hindi, Spanish, and Thai. The model is optimized for tasks requiring natural language understanding and generation, including complex reasoning, mathematical problem-solving, and code execution. Notably, it features a context window of 128,000 tokens, enabling it to handle extensive context for tasks like long-form text summarization and coding assistance.

For textual data, we utilize pre-trained language models to obtain contextualized embeddings for each token within a sequence. To derive a fixed-length representation of the entire sequence, we apply mean pooling. This involves calculating the average of all token embeddings, resulting in a comprehensive vector that encapsulates the semantic essence of the text.

\subsection{Visual Attribute Enconder}
For the visual attribute modality, we employ three pre-trained models to extract visual embeddings of the nodes: CLIP~\cite{CLIP}, ImageNet-Swinv2~\cite{Swinv2}, and FCMAE-ConvNeXTV2~\cite{convnextv2}.

\textbf{CLIP} (Contrastive Language-Image Pre-training) is a multimodal model developed by OpenAI that aligns images and text in a shared semantic space. It comprises an image encoder and a text encoder, both based on the Transformer architecture, trained jointly on a vast dataset of image-text pairs. This training enables CLIP to perform various tasks, such as zero-shot image classification and image-text retrieval, without requiring task-specific fine-tuning. We only use the image feature extractor (Vision Transformer) in CLIP.

\textbf{ImageNet-Swinv2} is the advanced version of the classical Swin Transformer, which is designed to handle large-scale high-resolution image data. It builds hierarchical feature maps by merging image patches in deeper layers, whose key feature lies in the linear computation complexity with respect to input image size, achieved by computing self-attention only within each local window.

\textbf{FCMAE-ConvNeXTV2} is a model that combines Fully Convolutional Masked Autoencoders (FCMAE) with the ConvNeXtV2 architecture. ConvNeXtV2 is an updated version of ConvNeXt, incorporating design elements from modern architectures to enhance performance. ConvNeXt-V2 is pretrained with a fully convolutional masked autoencoder framework (FCMAE) and fine-tuned on ImageNet-22k and then ImageNet-1k. We use the 88.7M params version of it which is uploaded in huggingface.

To extract visual features using the pre-trained vision models, the process begins by removing the classification head from the model.
This step is crucial as it allows us to retain only the feature extraction layers, which are responsible for capturing the underlying visual embeddings.

\subsection{Multimodal Attribute Enconder}
To effectively capture the multimodal attributes of nodes, we utilize two advanced open-source multimodal large language models available on Hugging Face: \textit{Llama3.2-11B-Vision-Instruct} and \textit{Qwen2-VL-7B-Instruct} \cite{qwen2}.

\textbf{Llama3.2-11B-Vision-Instruct} is part of Meta's Llama 3.2 series, introduced in September 2024. This model comprises 11 billion parameters and is instruction-tuned to enhance performance in tasks involving visual recognition, image reasoning, captioning, and answering general questions about images. Built upon the Llama 3.1 text-only model, it incorporates a vision adapter to process both text and image inputs, enabling a comprehensive understanding of multimodal data. The model supports a context length of up to 128,000 tokens, facilitating the handling of extensive contextual information.

\textbf{
Qwen2-VL-7B-Instruct} is a multimodal model developed by Qwen, featuring 7 billion parameters. This model is designed to process and generate both text and visual data, making it adept at tasks such as image captioning, visual question answering, and multimodal content generation. It employs a Naive Dynamic Resolution approach, allowing it to handle images of arbitrary resolutions by mapping them into a dynamic number of visual tokens. Additionally, it utilizes Multimodal Rotary Position Embedding (M-ROPE) to capture positional information across different modalities, enhancing its capability to understand complex multimodal inputs.

We input the textual attributes of the nodes along with the visual attributes through the processing pipeline of the VLM. 
Subsequently, we perform mean pooling on the image tokens and textual tokens from the last layer of hidden states in the LLM component of the VLM to obtain the final multimodal embeddings.

\section{Dataset}\label{Appendix-B}

\subsection{Dataset Documentation}

We standardize the data storage to facilitate the researchers' use of the datasets.
For each dataset, we have the following kinds of files or folders: 
\begin{itemize}
    \item "\textit{.csv}" format file, which holds the raw textual attribute information of the nodes.
    \item "\textit{.pt}" format file, which holds the graph data loaded in DGL~\cite{DGL}, including adjacency matrix and node labels. 
    \item "\textit{.gz}" formt file, which holds the image files named after the node ids (the image files are normally JPG or PNG).
    \item "\textit{Visutal Feature}" folder holds various image features in \textit{npy} format generated by PVMs.
    \item "\textit{Textual Feature}" folder holds various textual features in \textit{npy} format generated by PLMs. 
\end{itemize}

\subsection{Dataset Construction}\label{DatasetConstruction}
The construction of the multimodal attributed graph datasets includes the following three steps. 

First, preprocess the text and image information in the original data. For text, this includes removing missing values, non-English statements, abnormal symbols, length truncation, etc.  For images, entities that do not contain image links are removed.

Second, building the graph. The datasets constructed in MAGB are mainly from e-commerce data and social media data. 
For e-commerce data, Amazon already provides linking relationships between nodes, such as "also-view" and "also-buy" information between products (indicating the two product ids that are jointly purchased or viewed).
For social media data, we link posts by author-related information on the post, i.e., posts are linked if the same user has commented on both posts.
Note that self-edges and isolated nodes need to be removed when obtaining the final graph data. 

Third, refining the constructed graph. The graph nodes need corresponding numerical node labels for the node classification task. 
We convert the categories of nodes in the original data to numerical node labels in the graph. 
Detailed dataset class information will be shown later.

Fourth, we refine the image attributes. Based on the image URL, we download the corresponding image files and name each image with the associated node ID.
For nodes with multiple URLs, we only download the first image file.
We retrieve the relevant images from the Internet for failed image URLs based on their own textual attributes.
All downloaded images (including newly augmented images based on the retrieval) are guaranteed to be available for research and are not infringing.

\begin{figure}[t]
\vspace{-0.4cm}
\centering
\includegraphics[width=0.5\textwidth]{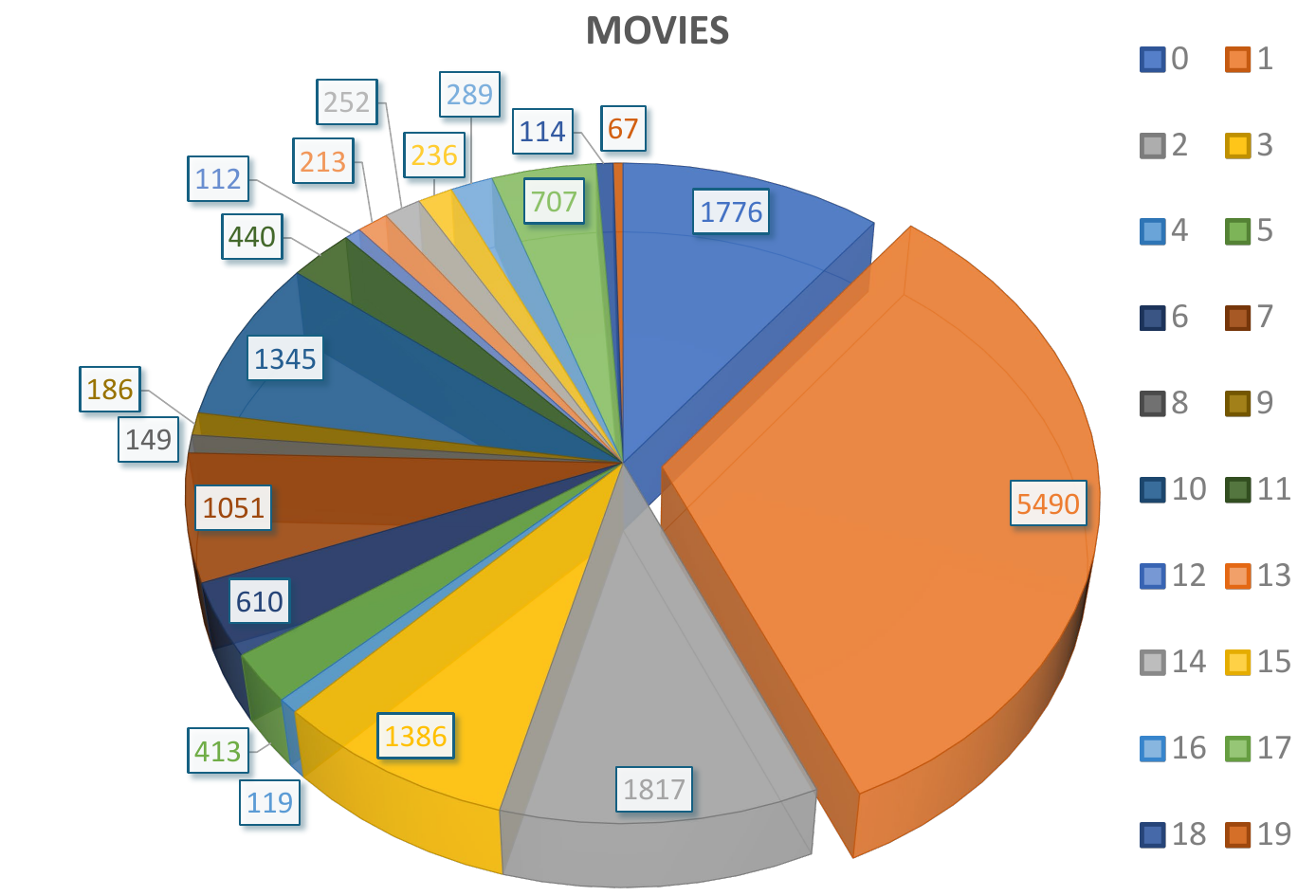}
\caption{Distribution of node labels in the Movies.}
\label{fig:movies label show}
\vspace{-0.3cm}
\end{figure}

\textbf{Movies} datasets are extracted from the "Movies \& TV" data in Amazon2018~\cite{Amazon}.
The original data can be found \href{https://datarepo.eng.ucsd.edu/mcauley_group/data/amazon_v2/metaFiles2/meta_Movies_and_TV.json.gz}{\textcolor{blue}{here}}.
The Movies dataset consists of items with the first-level label "Movies \& TV".
The nodes in the dataset represent Movies \& TV items on Amazon, and the edges mean two items are frequently co-purchased or co-viewed. 
The label of each dataset is the second-level label of the item.
We choose the title and description of the item itself as the text attributes of the node. 
We derive the visual attribute of the item based on the "imageURL" in the metadata.
The task is to classify the items into 20 categories.
The distribution of node labels is shown in Figure \ref{fig:movies label show}.

\begin{figure*}
	\centering
        \includegraphics[width=1.0\textwidth]{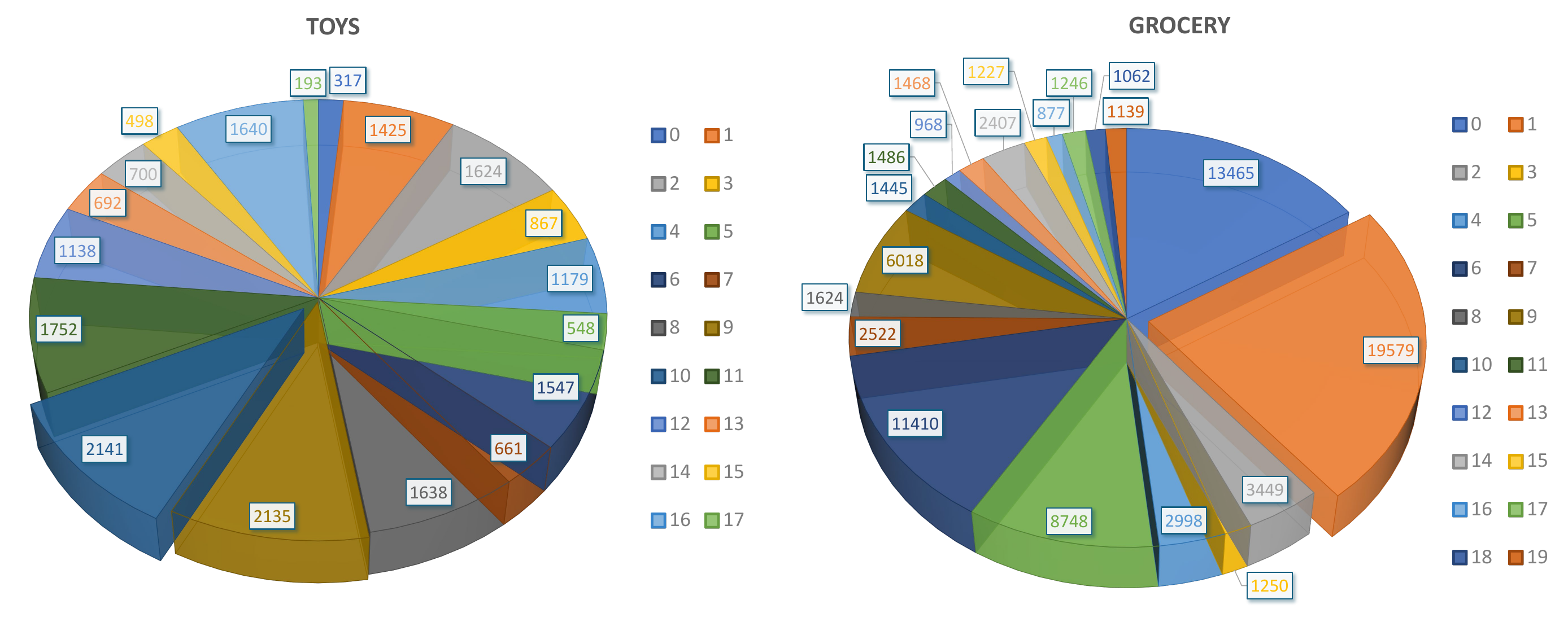}
	\caption{Distribution of node labels in the Toys and Grocery dataset.}
	\label{fig:Toys and Grocery show label}
 \vspace{-0.4cm} 
\end{figure*}

\begin{figure*}[h]
	\centering
        \includegraphics[width=1.0\textwidth]{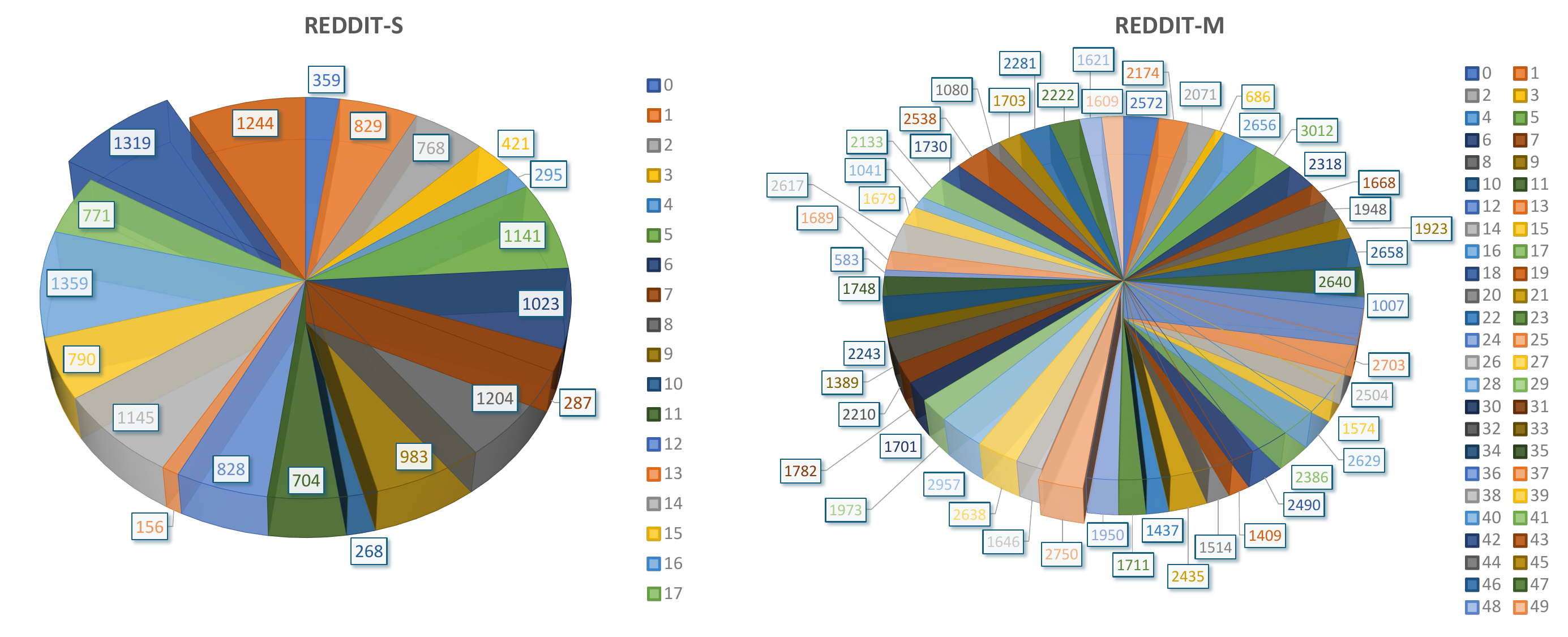}
	\caption{Distribution of node labels in the Reddit-S and Reddit-M dataset.}
	\label{fig:Reddit show label}
 \vspace{-0.4cm} 
\end{figure*}

\textbf{Toys} datasets are extracted from the "Toys \& Games" data in Amazon2018.
The original data can be found \href{https://datarepo.eng.ucsd.edu/mcauley_group/data/amazon_v2/metaFiles2/meta_Toys_and_Games.json.gz}{\textcolor{blue}{here}}.
The Toys dataset consists of items with the first-level label "Toys \& Games".
The nodes in the dataset represent toys or games items on Amazon, and the edges mean two items are frequently co-purchased or co-viewed. 
The construction of text attributes, visual attributes, and labels in the Toys dataset is consistent with Movies.
The task is to classify the items into 18 categories.
The distribution of node labels is shown in Figure \ref{fig:Toys and Grocery show label}.

\textbf{Grocery} datasets are extracted from the "Grocery \& Gourmet Food" data in Amazon2018.
The original data can be found \href{https://datarepo.eng.ucsd.edu/mcauley_group/data/amazon_v2/metaFiles2/meta_Grocery_and_Gourmet_Food.json.gz}{\textcolor{blue}{here}}.
The Grocery dataset consists of items with the first-level label "Grocery \& Gourmet Food".
The nodes in the dataset represent grocery or gourmet food items on Amazon, and the edges mean two items are frequently co-purchased or co-viewed. 
The construction of text attributes, visual attributes, and labels in the Grocery dataset is consistent with Movies.
The task is to classify the items into 20 categories.
The distribution of node labels is shown in Figure \ref{fig:Toys and Grocery show label}.

\textbf{Reddit-S/M} datasets are extracted from the RedCaps dataset~\cite{redcaps}.
The original data and a more detailed description of the data sources can be found \href{https://redcaps.xyz/download}{\textcolor{blue}{here}}.
The nodes in Reddit-S/M represent posts, which are linked if the same user has commented on both posts.
The node's visual attributes come from the image in the post, and the textual attributes come from the caption associated with the image.
The task on the two datasets is to classify posts into 20 and 50 categories, respectively.
The distribution of node labels in the two datasets is shown in Figure \ref{fig:Reddit show label}.


\section{Experimental Setup}
\subsection{Hardware Information}
All experiments are conducted on \textit{GPU}: NVIDIA V100 Tensor Core GPU 32GB using python 3.8, Transformers 4.39.2, dgl 0.8.0, and Pytorch 2.0,1.
A detailed and complete environment configuration can be found in the \href{https://github.com/sktsherlock/MAGB}{\textcolor{blue}{repository}}.

\begin{table*}[t]
  \centering
  \captionsetup{skip=5pt} 
  \caption{The node classification experimental results of TE-GNN and VE-GNN on the Grocery and Reddit-S datasets. Best results are highlighted with \colorbox[HTML]{52B788}{\textbf{bold}}, the second are marked with \colorbox[HTML]{95D5B2}{\underline{underline}}, and the third highest results are \colorbox[HTML]{D8F3DC}{marked}.}
    \begin{tabular}{clcccccccccc}
    \hline
    \multirow{2}[2]{*}{Settings} & \multicolumn{1}{c}{\multirow{2}[2]{*}{Feature}} & \multicolumn{5}{c}{Text} & \multicolumn{5}{c}{Visual}\\
      &   & \multicolumn{1}{l}{PLM} & \multicolumn{1}{l}{LLaMA1B} & \multicolumn{1}{l}{LLaMA8B} & \multicolumn{1}{l}{QwenVL} & \multicolumn{1}{l}{LLaMAVL} & \multicolumn{1}{l}{CLIP} & \multicolumn{1}{l}{ConvNeXT} & \multicolumn{1}{l}{Swin} & \multicolumn{1}{l}{QwenVL} & \multicolumn{1}{l}{LLaMAVL}\\
			\midrule
			\rowcolor{gray!8}\multicolumn{12}{c}{\textit{Dataset: Grocery. Metric: Accuracy}}\\
			\midrule
    \multicolumn{1}{c}{\multirow{4}[2]{*}{Full}} & GCN & 81.48 & 83.76 & \second{\underline{84.53}} & \third{84.42} & \first{\textbf{84.58}} & 79.87 & 78.18 & 78.56 & 81.76 & 81.08 \\
      & SAGE & 83.31 & 85.98 & \first{\textbf{86.65}} & \third{86.39} & \second{\underline{86.61}} & 81.11 & 79.34 & 79.73 & 83.19 & 82.46 \\
      & GAT & 80.26 & 82.08 & \second{\underline{83.05}} & \first{\textbf{83.09}} & \third{82.95} & 80.03 & 78.78 & 79.10 & 81.07 & 79.68 \\
      &  RevGAT & 83.58 & 86.46 & \second{\underline{87.21}} & \third{86.97} & \first{\textbf{87.41}} & 81.81 & 78.63 & 78.95 & 83.18 & 82.14 \\
    \hline
    \multicolumn{1}{c}{\multirow{4}[2]{*}{10-shot}} & GCN & 57.88 & 63.41 & \second{\underline{64.50}} & \third{63.67} & \first{\textbf{64.72}} & 59.29 & 54.01 & 53.02 & 61.76 & 61.20 \\
      & SAGE & 55.92 & \third{66.08} & \first{\textbf{68.01}} & 65.67 & \second{\underline{67.98}} & 60.70 & 52.23 & 50.39 & 61.23 & 62.89 \\
      & GAT & 56.98 & \third{61.25} & \second{\underline{62.11}} & 60.64 & \first{\textbf{62.17}} & 57.40 & 55.02 & 54.52 & 58.92 & 61.11 \\
      & RevGAT  & 52.98 & 64.11 & \first{\textbf{67.07}} & \third{64.76} & \second{\underline{66.98}} & 58.34 & 44.88 & 43.67 & 55.94 & 55.09 \\
    \hline
    \multicolumn{1}{c}{\multirow{4}[2]{*}{ 3-shot}} & GCN & 44.10 & \third{52.04} & \first{\textbf{52.57}} & 49.91 & \second{\underline{52.37}} & 46.27 & 39.47 & 40.04 & 46.46 & 48.57 \\
      & SAGE & 36.78 & \first{\textbf{47.29}} & \second{\underline{47.16}} & 46.08 & \third{47.11} & 46.60 & 38.08 & 37.91 & 45.35 & 46.84 \\
      & GAT & 41.47 & \third{47.64} & 46.80 & 46.13 & \second{\underline{48.02}} & 44.13 & 44.37 & 44.46 & 46.53 & \first{\textbf{49.35}} \\
      & RevGAT & 31.09 & \third{45.83} & \first{\textbf{48.41}} & 45.73 & \second{\underline{47.48}} & 41.62 & 32.36 & 27.68 & 39.77 & 35.00 \\
			\midrule
			\rowcolor{gray!8}\multicolumn{12}{c}{\textit{Dataset: Reddit-S. Metric: F1-Macro}}\\
			\midrule
    \multicolumn{1}{c}{\multirow{4}[2]{*}{Full}} & GCN & 88.41 & \first{\textbf{89.59}} & \second{\underline{89.58}} & 89.34 & \third{89.54} & 88.66 & 87.91 & 87.89 & 88.70 & 88.55 \\
      & SAGE & 88.84 & 90.61 & \second{\underline{90.72}} & \third{90.64} & \first{\textbf{90.79}} & 90.47 & 89.15 & 89.09 & 90.37 & 90.52 \\
      & GAT & 88.78 & \first{\textbf{89.53}} & \second{\underline{89.41}} & 89.21 & \third{89.40} & 88.43 & 87.59 & 87.53 & 88.47 & 88.36 \\
      & RevGAT & 88.84 & 90.54 & \first{\textbf{90.75}} & 90.53 & \second{\underline{90.65}} & 89.41 & 88.83 & 89.18 & 90.08 & \third{90.58} \\
    \hline
    \multicolumn{1}{c}{\multirow{4}[2]{*}{ 10-shot}} & GCN & 70.32 & 80.13 & 81.49 & 80.05 & 81.17 & \second{\underline{88.76}} & 86.15 & 86.69 & \third{88.06} & \first{\textbf{89.29}} \\
      & SAGE & 62.69 & 72.50 & 74.11 & 72.89 & 72.93 & \second{\underline{82.69}} & 79.74 & 79.99 & \third{81.97} & \first{\textbf{83.28}} \\
      & GAT & 68.51 & 74.03 & 75.88 & 69.73 & 73.20 & \first{\textbf{88.77}} & 85.13 & 84.43 & \third{86.34} & \second{\underline{87.73}} \\
      & RevGAT & 60.81 & 69.59 & 69.93 & 67.99 & 70.72 & \second{\underline{82.58}} & 79.73 & 79.53 & \third{81.50} & \first{\textbf{82.87}} \\
    \hline
    \multicolumn{1}{c}{\multirow{4}[2]{*}{ 3-shot}} & GCN & 58.67 & 69.65 & 72.93 & 72.86 & 72.62 & \first{\textbf{87.45}} & 84.13 & 83.41 & \third{85.89} & \second{\underline{86.16}} \\
      & SAGE & 44.49 & 54.04 & 59.80 & 61.42 & 60.30 & \first{\textbf{78.29}} & 74.78 & 74.16 & \third{76.63} & \second{\underline{77.91}} \\
      & GAT & 50.77 & 50.30 & 55.70 & 54.15 & 55.60 & \first{\textbf{85.10}} & 79.15 & 75.86 & \third{80.64} & \second{\underline{82.41}} \\
      & RevGAT & 46.95 & 53.92 & 55.55 & 53.55 & 56.75 & \first{\textbf{76.87}} & 73.12 & 71.20 & \third{74.04} & \second{\underline{75.65}} \\
    \hline
    \end{tabular}%
  \label{tab:NCofGrocery}%
\end{table*}%
\begin{table*}[ht]
  \centering
   \captionsetup{skip=5pt} 
  \caption{Comparison of ME-GNN with TE-GNN and VE-GNN on the node classification task under different settings.}
  \begin{adjustbox}{width=0.94\textwidth}
    \begin{tabular}{clcccc|cccc|cccc}
    \hline
    \multirow{2}[2]{*}{Datasets} & \multicolumn{1}{c}{\multirow{2}[2]{*}{Feature}} & \multicolumn{4}{c|}{Full} & \multicolumn{4}{c|}{10-Shot} & \multicolumn{4}{c}{3-Shot} \\
      &   & GCN & SAGE & GAT & RevGAT & GCN & SAGE & GAT & RevGAT & GCN & SAGE & GAT & RevGAT \\
    \hline
    \multirow{5}[2]{*}{Grocery} & Text & {\underline{84.58}} & 86.65 & {\underline{83.09}} & {\textbf{87.41}} & 64.72 & {\underline{68.01}} & {\underline{62.17}} & {\underline{67.07}} & {\underline{52.57}} & 47.29 & 48.02 & 48.41 \\
      & Visual & 81.76 & 83.19 & 81.07 & 83.18 & 61.76 & 62.89 & 61.11 & 58.34 & 48.57 & 46.84 & {\textbf{49.35}} & 41.62 \\
      & LLaMA+CLIP & {\textbf{84.64}} & {\underline{86.65}} & {\textbf{83.17}} & {\underline{87.40}} & {\underline{64.84}} & 67.91 & 61.24 & 66.95 & 52.55 & 48.08 & 46.34 & {\underline{48.79}} \\
      & QwenVL & 82.96 & 85.55 & 82.10 & 85.77 & 63.90 & 65.96 & 60.11 & 62.43 & 50.02 & {\underline{51.05}} & 46.98 & 45.54 \\
      & LLaMAVL & 84.17 & {\textbf{86.81}} & 82.88 & 87.39 & {\textbf{65.63}} & {\textbf{69.95}} & {\textbf{62.55}} & {\textbf{68.91}} & {\textbf{53.83}} & {\textbf{51.88}} & {\underline{48.80}} & {\textbf{50.20}} \\
    \hline
    \multirow{5}[2]{*}{Reddit-S} & Text & 89.59 & 90.79 & 89.53 & 90.75 & 81.49 & 74.11 & 75.88 & 70.72 & 72.93 & 61.42 & 55.70 & 56.75 \\
      & Visual & 88.70 & 90.52 & 88.47 & 90.58 & {\underline{89.29}} & {\underline{83.28}} & {\underline{88.77}} & {\underline{82.87}} & {\textbf{87.45}} & {\textbf{78.29}} & {\textbf{85.10}} & {\textbf{76.87}} \\
      & LLaMA+CLIP & 89.77 & 91.08 & 89.73 & 91.01 & 81.85 & 74.42 & 74.37 & 70.87 & 73.11 & 61.04 & 58.13 & 56.66 \\
      & QwenVL & {\underline{90.88}} & {\underline{92.57}} & {\underline{90.49}}& {\underline{92.39}} & 88.66 & 82.76 & 86.14 & 81.62 & 86.24 & 76.67 & 80.33 & 74.30 \\
      & LLaMAVL & {\textbf{91.13}} & {\textbf{92.94}} & {\textbf{90.77}} & {\textbf{92.85}} & {\textbf{90.50}} & {\textbf{84.92}} & {\textbf{88.85}} & {\textbf{84.27}} & {\underline{87.35}} & {\underline{77.19}} & {\underline{80.59}} & {\underline{75.92}} \\
    \hline
    \end{tabular}%
  \label{tab:groceryandreddits}%
  \end{adjustbox}
  \vspace{-5pt}
\end{table*}%
\begin{table*}[htbp]
  \centering
  \captionsetup{skip=5pt} 
  \caption{Comparison of ME-GNN with TE-GNN and VE-GNN on the link prediction task}
  \begin{adjustbox}{width=0.94\textwidth}
    \begin{tabular}{lccc|ccc|ccc|ccc|ccc}
    \hline
    \multicolumn{1}{c}{\multirow{2}[2]{*}{Feature}} & \multicolumn{3}{c}{Movies} & \multicolumn{3}{c}{Toys} & \multicolumn{3}{c}{Grocery} & \multicolumn{3}{c}{Reddit-S} & \multicolumn{3}{c}{Reddit-M} \\
      & Hits1 & Hits3 & \multicolumn{1}{c}{MRR} & Hits1 & Hits3 & \multicolumn{1}{c}{MRR} & Hits1 & Hits3 & \multicolumn{1}{c}{MRR} & Hits1 & Hits3 & \multicolumn{1}{c}{MRR} & Hits1 & Hits3 & MRR \\
			\midrule
			\rowcolor{gray!8}\multicolumn{16}{c}{\textit{Backbone model: GraphSAGE}}\\
			\midrule
 RoBERTa & 10.33 & 28.96 & 25.95 & 12.28 & 32.74 & 28.06 & 10.56 & 29.53 & 25.89 & \first{\textbf{8.34}} & \first{\textbf{22.27}} & \first{\textbf{21.33}} & \first{\textbf{66.82}} & 87.81 & \first{\textbf{78.85}} \\
           LLaMA1B & 10.15 & 28.44 & 26.00 & 11.44 & 32.22 & 27.68 & 9.74  & 27.70 & 24.86 & \second{\underline{8.02}}  & 21.95 & \second{\underline{21.18}} & 66.42 & \second{\underline{88.00}} & 78.66 \\
          LLaMA8B & 10.46 & 30.05 & 26.69 & 12.44 & 33.05 & 28.67 & 9.99  & 28.86 & 25.72 & 7.89  & 21.88 & 21.02 & 66.36 & 87.78 & 78.67 \\
          QWenVL-T & 10.29 & 29.98 & 26.42 & 12.16 & 34.62 & 29.46 & 10.43 & 29.88 & 26.22 & 7.91  & 21.87 & 21.17 & 65.96 & 87.81 & 78.50 \\
          LLaMAVL-T & 10.42 & 30.03 & 26.83 & 12.16 & 33.48 & 28.20 & 10.41 & 29.23 & 26.21 & 7.76  & 21.87 & 21.17 & \second{\underline{66.58}} & 87.80 & \second{\underline{78.68}} \\
          \hline
          CLIP  & \second{\underline{10.85}} & \first{\textbf{31.30}} & \first{\textbf{27.65}} & 12.26 & 33.38 & 28.55 & 10.65 & 30.32 & 26.36 & 7.97  & 21.62 & 20.96 & 66.31 & 87.26 & 78.31 \\
          ConvNeXT & \first{\textbf{11.06}} & 29.81 & 27.02 & \second{\underline{13.23}} & \second{\underline{36.53}} & \second{\underline{30.56}} & \first{\textbf{11.03}} & \first{\textbf{31.20}} & \second{\underline{27.06}} & 7.58  & 21.52 & 20.76 & 65.83 & 87.29 & 78.07 \\
          SwinV2 & 10.74 & \second{\underline{30.64}} & \second{\underline{27.32}} & \first{\textbf{13.36}} & \first{\textbf{37.61}} & \first{\textbf{31.36}} & \second{\underline{10.92}} & \second{\underline{30.83}} & \first{\textbf{27.17}} & 7.61  & 21.32 & 20.75 & 65.78 & 87.18 & 78.03 \\
          QWenVL-V & 10.26 & 29.18 & 26.39 & 11.91 & 32.58 & 28.24 & 9.55  & 28.49 & 25.20 & 7.26  & 20.93 & 20.34 & 65.62 & 87.02 & 77.89 \\
          LLaMAVL-V & 8.88  & 24.85 & 23.26 & 9.32  & 26.01 & 23.27 & 8.20  & 22.89 & 21.17 & 6.52  & 19.39 & 19.30 & 55.83 & 83.53 & 71.50 \\
          \hline
          LLaMA+CLIP & 10.59 & 30.42 & 27.03 & 11.94 & 34.17 & 28.99 & 9.81  & 29.81 & 25.89 & 7.86  & \second{\underline{21.99}} & 20.92 &  66.41     &  87.83     & 78.57 \\
          QWenVL & 10.42 & 29.57 & 26.49 & 12.52 & 34.29 & 29.22 & 10.66 & 29.49 & 26.28 & 7.91  & 21.54 & 21.03 & 66.30 & \first{\textbf{88.00}} & 78.64 \\
          LLaMAVL & 10.70 & 29.89 & 26.91 & 11.89 & 33.34 & 28.48 & 10.35 & 28.68 & 26.25 & 7.80  & 21.91 & 21.09 & 66.30 & 87.80 & 78.65 \\
      \hline
    \end{tabular}%
  \label{tab:linksage}%
  \end{adjustbox}
\end{table*}%

\subsection{Experimental Setting of GNNs}\label{app:spliting}
We conduct experiments on 4 GNN models described in A.1 on 5 datasets.
We use the aforementioned attribute encoder to model the node attributes and form the initial node features of the graph data. 
Each experiment is repeated ten times, and the evaluation metrics are Accuracy and the F1-Macro score. 
The parameters shared by all GNN models include epochs, model layers, hidden units, label smoothing factor, learning rate, and dropout ratio, and their values are set to 1000, \{2,3\}, \{64,128,256\}, 0.1, \{5e-03, 5e-02, 1e-02, 2e-02\}, \{0.2, 0.5, 0.75\}, respectively.
In addition to these hyperparameter settings shared above, for GAT and RevGAT, we freeze the number of heads to 3 and set the ratio of attention-dropout and edge-dropout to 0 and 0.25 by default.
For the GraphSAGE model, we use the mean pool to aggregate the neighbor information.
The eval patience of all models is set to 1.
We use cross-entropy loss with the AdamW optimizer to train and optimize all the above models. 
GNNs are mainly derived from the implementation in the DGL library~\cite{DGL}.

For dataset splitting, we consider both the supervised learning scenario and two few-shot learning settings.
In the supervised learning scenario, each dataset is split into 60\% training, 20\% validation, and 20\% test sets. 
In contrast, for the k-shot learning scenario(e.g., $k=3,10$), we randomly select k instances per class to form the training set, while the remaining samples are split into 20\% validation and 80\% test sets. 
These different data partitioning strategies allow us to evaluate model performance under both extensive supervision and low-data scenarios, providing a more comprehensive assessment of GNNs in multimodal graph learning.

For the link prediction experiment, the number of GNN layers, hidden layer dimensions, learning rate, dropout rate, and batch size are set to 3, \{128, 256\}, 0.01, 0.02, and 2048, respectively. For most datasets, we set the number of negative samples to 5000. However, due to the larger size of the Reddit-M dataset, we set the number of negative samples to 2000 to reduce model evaluation time.

\section{Additional Experimental Results}\label{app:experiments}

\subsection{Node Classification}
Table {\color{brown}\ref{tab:NCofGrocery}-\ref{tab:groceryandreddits}} shows the node classification results on the remaining two datasets, Grocery and Reddit-S. 
Observing from Table {\color{brown}\ref{tab:NCofGrocery}}, we can find that for the e-commerce network Grocery, textual features consistently achieve top-three performance across almost all experimental settings.
In contrast, for the social network Reddit-S, under the more challenging few-shot learning conditions, visual features exhibit a significant advantage.
This finding is largely consistent with the conclusion drawn in Section \ref{sec:3.2}.

Table {\color{brown}\ref{tab:groceryandreddits}} presents the impact of multimodal embeddings on GNN performance. As shown, on the Reddit-S dataset, the multimodal embeddings generated by LLaMA-VL outperform those produced by the combination of LLaMA-8B and CLIP, with average improvements of 1.53\%, 11.76\%, and 18.03\% across the three learning scenarios. This suggests that in few-shot learning settings, modality bias in multimodal embeddings may hinder downstream GNNs from effectively leveraging informative modal attributes. In contrast, VLM-generated multimodal embeddings help mitigate this issue, aligning with our analysis in the main text.

\begin{figure*}[t]
  \centering
  \setlength{\abovecaptionskip}{5pt} 
  \includegraphics[width=0.98\textwidth]{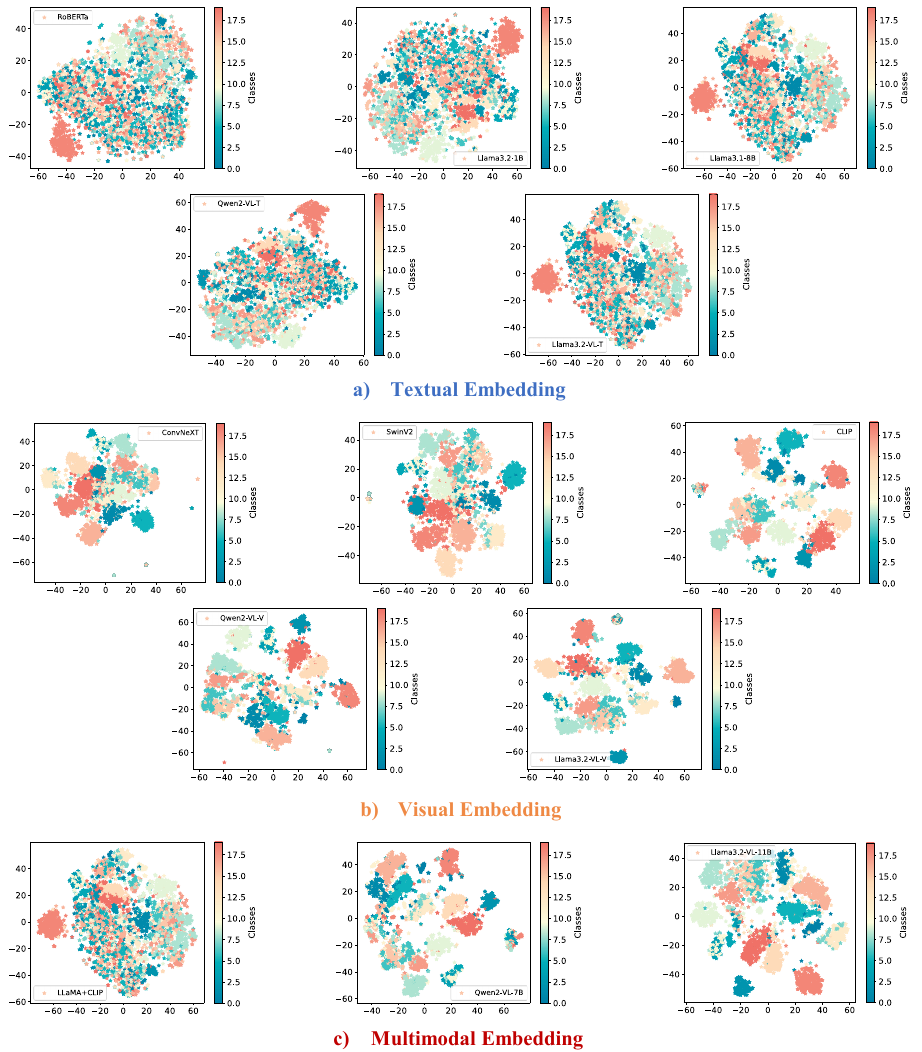}
  \caption{TSNE visualization of different embeddings on Reddit-S dataset.}
    \vspace{-10pt} 
  \label{fig:tsne-reddit-s}
\end{figure*}

In Figure {\color{brown}\ref{fig:tsne-reddit-s}}, we present the t-SNE visualizations of all modal representations on the Reddit-S dataset. From top to bottom, the plots correspond to textual embeddings, visual embeddings, and multimodal embeddings, respectively. It is evident that the spatial distribution of the ‘LLaMA+CLIP’ multimodal representation, which is formed by concatenating the LLaMA textual embeddings with the CLIP visual embeddings, closely aligns with the text embeddings from \textit{LLaMA3.1-8B}. This indicates that the lower-dimensional yet better-clustered CLIP visual features are almost disregarded. Furthermore, this observation highlights that simply combining representations from different modalities without proper alignment can lead to significant consequences.



\subsection{Link Prediction}
Table {\color{brown}\ref{tab:linksage}} presents the performance of various modality embeddings on the link prediction task using GraphSAGE as the backbone. 
Notably, ConvNeXT, which performed well with GCN, also achieves strong results with GraphSAGE. 
Compared to text representations generated by generative large language models, those produced by the traditional pretrained language model RoBERTa appear to be more suitable for downstream link prediction tasks. Specifically, in the evaluations on the RedditS/M datasets, RoBERTa consistently achieves the best performance in most cases.
Furthermore, for e-commerce network datasets, RoBERTa outperforms the larger LLaMA1B model in terms of Hits@1, Hits@3, and MRR across the three datasets, with average improvements of 0.61\%, 0.96\%, and 0.45\%, respectively.
These phenomena further indicate that different features may exhibit varying effectiveness across different downstream tasks.
Furthermore, the relatively minor performance differences observed across features, datasets, and evaluation metrics underscore that graph topology plays a dominant role in link prediction tasks.
This suggests that for link prediction tasks, graph topology is of paramount importance. Enhancing structural modeling is likely to yield more significant performance gains than focusing solely on the refinement of node attribute representations. 

\subsection{Uncontrollable Output Analysis.}
\label{app:uncontroll}

\begin{figure}[t]
  \centering
  \captionsetup{aboveskip=2pt, belowskip=3pt} 
  \includegraphics[width=0.46\textwidth]{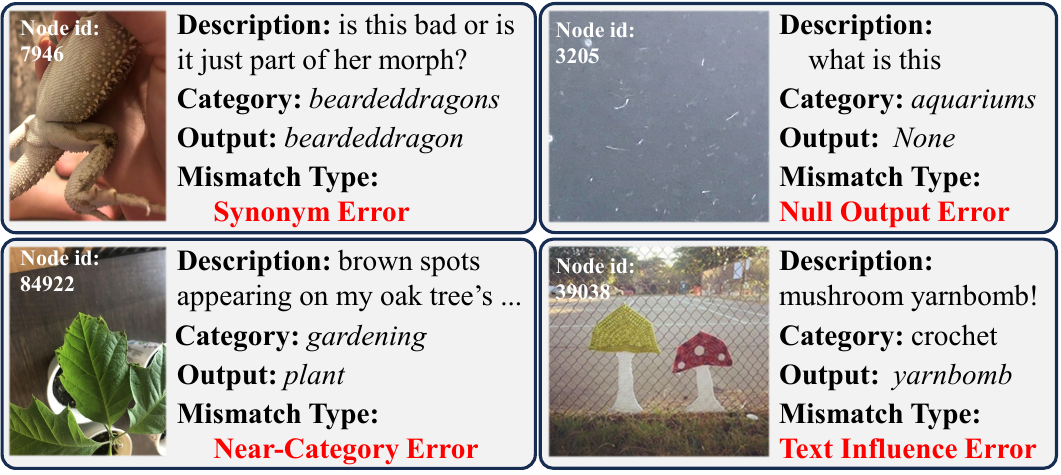}
  \caption{VLM Inference Mismatch Types.}
    \vspace{-10pt} 
  \label{fig:VLM_case}
\end{figure}

To better understand the uncontrollable output issue of the \textit{Qwen2-VL} model, we categorize its misclassifications into four distinct types and present representative cases in Figure~\ref{fig:VLM_case}. The identified error types are as follows:  

\begin{itemize}
    \item \textbf{Synonym Error:} The model predicts a label that is a close synonym of the ground truth but does not match exactly. As shown in the top-left of Figure \ref{fig:VLM_case}, when the true category is \textit{“beardeddragon”}, the model outputs \textit{“beardeddragon”}. 
    \item \textbf{Near-Category Error:} The model misclassifies an instance into a closely related category instead of the ground truth. As depicted in the bottom-left of Figure \ref{fig:VLM_case}, the correct label \textit{“gardening”} is mistakenly predicted as \textit{“plant”}, while the predefined category list includes \textit{“plants”}.
    \item \textbf{Null Output Error:} In some cases, the model fails to generate any meaningful output, likely due to insufficient or ambiguous information in the provided image and text. This phenomenon is illustrated in the top-right of Figure  \ref{fig:VLM_case}. 
    \item \textbf{Text Influence Error:} The model’s prediction is affected by textual attributes associated with the image, leading to an incorrect output. In the bottom-right of Figure  \ref{fig:VLM_case}, when the true category is missing from the generated label list, the model outputs \textit{“yarnbomb”} instead of a valid category, influenced by its associated text \textit{“mushroom yarnbomb”}. 
\end{itemize}

These observations highlight the challenges posed by uncontrolled generation in vision-language models, emphasizing the need for improved alignment techniques to mitigate such errors.

\subsection{Additional Observations}
Due to the space of the main text, we supplement some experimental conclusions here.

\textbf{Generative Large Vision-Language models can serve as powerful visual encoders.} We observe that compared to traditional pre-trained vision models, the visual representations obtained from vision-language models achieve average improvements of 2.08\%, 3.23\%, and 3.51\% across the three learning scenarios on the Movies dataset. Similarly, on the Toys dataset, the average improvements across the three learning scenarios are 1.17\%, 2.44\%, and 6.76\%, respectively.

\textbf{Different large vision-language models exhibit varying strengths in multimodal embeddings extraction.} 
Observing Table {\color{brown}\ref{tab:NC of MEGNN}}, we notice that although the total parameter count of the Qwen2VL and LLaMA Vision models differs by approximately 4B, the multimodal embeddings extracted by the Qwen model are comparable and even superior to those extracted by LLaMA on the e-commerce network dataset. For instance, on the Movies dataset, QwenVL achieves average improvements of 0.10\%, 2.56\%, and 1.28\% over LLaMAVL in supervised learning, 10-shot, and 3-shot scenarios, respectively. In contrast, LLaMA performs better on the social network dataset Reddit-M, with average improvements of 3.73\%, 4.19\%, and 2.11\%.
This performance discrepancy is likely closely related to the distribution of the pretraining data of the models. On the one hand, since Qwen was developed by Alibaba in China, it may have been exposed to a large number of product images and descriptions from e-commerce platforms during pretraining, which explains its superior performance on datasets related to e-commerce scenarios. On the other hand, the pretraining data of LLaMA probably covers more content from social networking platforms, endowing it with stronger adaptability on datasets related to social networks like Reddit-M. Therefore, this performance difference not only reflects the design advantages of the model architectures but also highlights the significant impact of pretraining data on multimodal representation capabilities.

\subsection{Case Study of Zero-shot Learning}
\begin{figure}[ht]
  \centering
  \captionsetup{aboveskip=2pt, belowskip=0pt} 
  \includegraphics[width=0.47\textwidth]{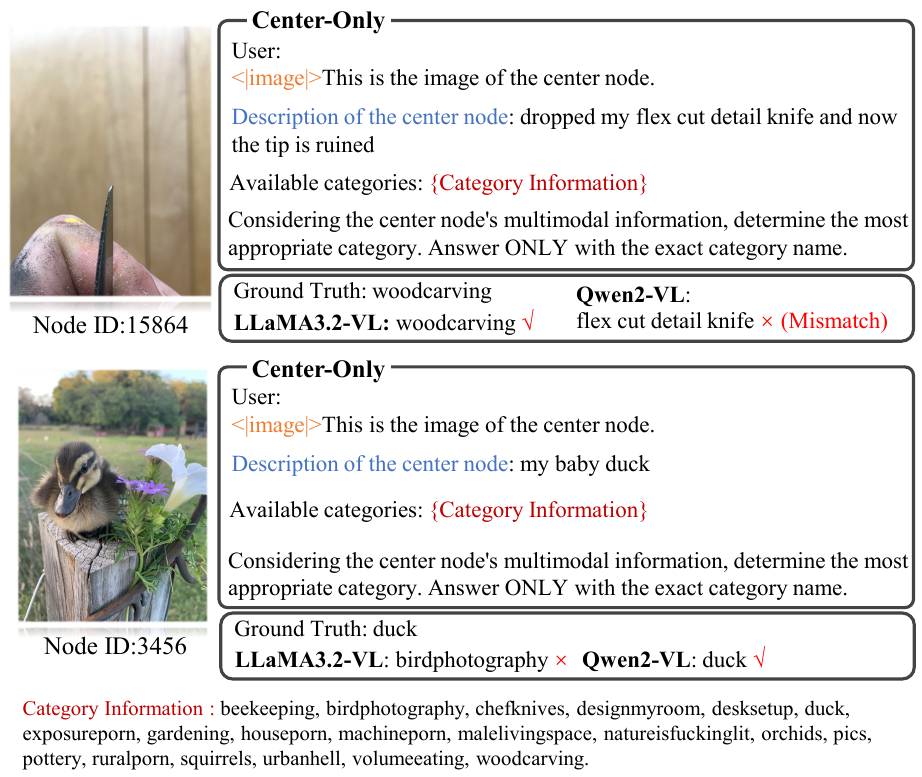}
  \caption{Case study of zero-shot learning.}
    \vspace{-10pt} 
  \label{fig:zero-shot}
\end{figure}

In this subsection, we present case studies on zero-shot node classification using \textit{LLaMA3.2-11B-Vision} and \textit{Qwen2-VL-7B} on the Reddit-S datasets. 
As shown in Figure \ref{fig:zero-shot}, one can be observed that LLaMA-VL's output is generally more controllable in most cases. 
Moreover, when misclassifications occur, the predicted categories often remain semantically close to the ground-truth labels. For instance, misclassifying "duck" as "bird photography" still reflects a reasonable semantic relationship, indicating that LLaMA-VL effectively captures contextual similarities even in cases of prediction errors. This suggests that LLaMA-VL's predictions preserve meaningful semantic structures, making its outputs more interpretable and aligned with human perception.

\subsection{Case Study of Graph Retrieval}
\begin{figure}[ht]
  \centering
  \captionsetup{aboveskip=2pt, belowskip=0pt} 
  \includegraphics[width=0.48\textwidth]{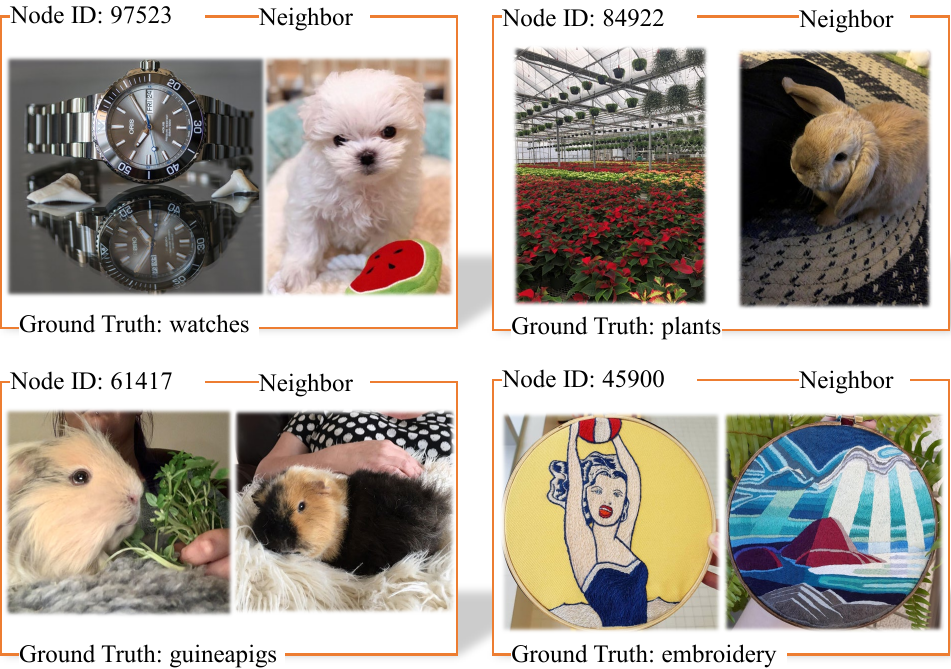}
  \caption{Case study of Graph Retrieval in Reddit-M Dataset.}
    \vspace{-10pt} 
  \label{fig:GRE_case_REDDIT}
\end{figure}

In this subsection, we visualize the image information retrieved from the graph structure during the experiments.
Observing Figure {\color{brown}\ref{fig:GRE_case_REDDIT}}, we notice that the original image of node 97523 in Reddit-M is a watch, and the category of its corresponding post is also "watches." 
However, the image retrieved from its first-order neighbors is of a small dog. 
The introduction of such information, which significantly differs from the original category, can negatively impact the classification task of the original node. 
A similar observation is made for node 84922, where the original image depicts a flowerbed, and its corresponding category is "plants," yet the retrieved neighbor image is of a rabbit.

Below Figure  {\color{brown}\ref{fig:GRE_case_REDDIT}}, we show cases where neighbor information positively enhances the original node's classification.
In summary, when employing retrieval-augmented techniques on MAGs, it may be necessary to adopt practices from graph structure learning to ensure that the retrieved neighbors contribute positively to downstream tasks.

\subsection{Future Direction}
\label{app:future_direction}
Based on the extensive experiments and analyses conducted above, we have summarized the following directions to further explore representation learning on MAGs and enhance the effectiveness of these methods in real-world applications.

\ding{172} \textbf{Exploring Better VLM Generation Strategies for Multi-modal Representations.}  
Our preliminary work demonstrates the potential of Vision-Language Models (VLMs) as multimodal representation extractors. Future research should explore more sophisticated generation strategies to obtain richer and more balanced multimodal representations. For instance, cross-modal alignment training techniques could be employed to reduce discrepancies between textual and visual modalities. Additionally, prompt-based methods and adaptive fine-tuning strategies may further enhance the quality and informativeness of the generated multimodal features, thereby leading to more robust downstream performance.

\ding{173} \textbf{Developing Multi-modal GNNs for Enhanced Representation Learning in MAGs.}  
Current Graph Neural Network (GNN) architectures are predominantly tailored for unimodal data, which limits their ability to fully leverage the diverse information contained within MAGs. An important future direction is the design of novel multimodal GNN architectures that seamlessly integrate and process heterogeneous data sources. This could involve the development of fusion layers that dynamically balance the contributions from different modalities or the incorporation of attention mechanisms to better capture inter-modal relationships. Such advancements would significantly improve the understanding of complex node interactions and boost performance in tasks like node classification, link prediction, and clustering.

\ding{174} \textbf{Exploring New Learning Paradigms for MAGs.}  
Innovative learning paradigms that integrate the strengths of both GNNs and VLMs represent a promising research avenue. Future work may focus on collaborative learning frameworks where GNNs and VLMs are jointly optimized to complement each other's abilities. For example, one could design co-training schemes or hybrid architectures that leverage the structural insights from GNNs and the rich semantic understanding from VLMs.

\ding{175} \textbf{Designing Effective Prompts for Integrating Neighboring Multi-modal Information}
With the rapid development of prompt-based learning, another compelling research direction is to design effective prompts that can integrate multimodal information from neighboring nodes in MAGs. Crafting context-aware prompts could help guide models to capture the nuanced relationships among nodes, thereby enhancing the overall representation quality. Future studies might investigate dynamic prompting strategies that adapt to the varying importance of local neighborhoods, ensuring that the combined information from multiple modalities is utilized in a balanced and context-sensitive manner.

\ding{176} \textbf{Researching Graph Retrieval Enhancement Strategies for Efficient Inference}
Graph Retrieval Enhancement methods offer a promising pathway to improve inference efficiency and model performance by integrating graph-based retrieval techniques. Future research could focus on developing retrieval strategies that dynamically fetch relevant multimodal samples during inference, thereby providing additional context that enhances prediction accuracy. The key challenge will be to design systems that balance relevance with computational efficiency, ensuring that the retrieved multimodal information seamlessly complements the model’s internal representations without overwhelming it.

\ding{177} \textbf{Enriching MAG Datasets for Broader Evaluation.}
To foster the development of robust multimodal representation learning methods, there is a pressing need to expand and diversify MAG datasets. Future efforts should consider:
\begin{enumerate} \item \textbf{Domain Diversity:} Incorporating data from a wider range of domains such as healthcare, finance, and autonomous driving can significantly enhance the model's adaptability and generalization across different application scenarios. \item \textbf{Modality Expansion:} Integrating additional modalities, including audio, video, and time-series data, will further enrich the available information and enable the exploration of more complex multimodal interactions. \item \textbf{Dynamic Graph Data:} Constructing dynamic MAG datasets that capture the temporal evolution of both graph structures and multimodal attributes will promote research into time-aware multimodal representation learning, enabling models to better understand and predict time-varying interactions. \end{enumerate}

These directions collectively outline a roadmap for future research, aiming to bridge the current gaps in multimodal graph representation learning and ultimately lead to more effective, scalable, and interpretable models for real-world applications.

\end{document}